\documentclass[pdflatex]{sn-jnl}%
\usepackage{graphicx}%
\usepackage{multirow}%
\usepackage{amsmath,amssymb,amsfonts}%
\usepackage{amsthm}%
\usepackage{mathrsfs}%
\usepackage[title]{appendix}%
\usepackage{xcolor}%
\usepackage{textcomp}%
\usepackage{manyfoot}%
\usepackage{booktabs}%
\usepackage{makecell}
\usepackage{float}
\usepackage{threeparttable}
\usepackage{algorithm}%
\usepackage{algorithmicx}%
\usepackage{algpseudocode}%
\usepackage{listings}%

\usepackage{caption}%
\usepackage{subcaption}%
\theoremstyle{thmstyleone}%
\theoremstyle{thmstyletwo}%

\theoremstyle{thmstylethree}%

\begin{document}

\title[Article Title]{Temporal Fusion Nexus: A task-agnostic multi-modal embedding model for clinical narratives and irregular time series in post-kidney transplant care}

\author*[1,2]{\fnm{Aditya} \sur{Kumar}}

\author[2]{\fnm{Simon} \sur{Rauch}}

\author[1,2]{\fnm{Mario} \sur{Cypko}}

\author[3]{\fnm{Marcel} \sur{Naik}}

\author[4]{\fnm{Matthieu-P} \sur{Schapranow}}

\author[4]{\fnm{Aadil} \sur{ Rashid}}

\author[3]{\fnm{Fabian} \sur{ Halleck}}

\author[3]{\fnm{Bilgin} \sur{Osmanodja}}

\author[5]{\fnm{Roland} \sur{Roller}}

\author[6]{\fnm{Lars} \sur{Pape}}

\author[3]{\fnm{Klemens} \sur{Budde}}

\author[7]{\fnm{Mario} \sur{Schiffer}}

\author[1,2]{\fnm{Oliver} \sur{Amft}}

\affil[1]{\orgdiv{} \orgname{Hahn-Schickard}, \orgaddress{\street{} \city{Freiburg} \postcode{}, \country{Germany}}}

\affil[2]{\orgdiv{} Intelligent Embedded Systems Lab, \orgname{University of Freiburg}, \orgaddress{\street{} \city{Freiburg}, \postcode{} \country{Germany}}}

\affil[3]{\orgdiv{}\orgname{Charité University Medical Center}, \orgaddress{\street{} \city{Berlin}, 
%\postcode{}, 
\country{Germany}}}

\affil[4]{\orgdiv{}  \orgname{Hasso Plattner Institute for Digital Engineering, University of Potsdam}, \orgaddress{\street{} \city{Potsdam},
%\postcode{}, 
\country{Germany}}}

\affil[5]{\orgdiv{}  \orgname{DFKI}, \orgaddress{\street{} \city{Berlin},
%\postcode{}, 
\country{Germany}}}

\affil[6]{\orgdiv{}  \orgname{University Hospital Essen}, \orgaddress{\street{} \city{Essen},
%\postcode{}, 
\country{Germany}}}

\affil[7]{\orgdiv{} \orgname{University Hospital Erlangen}, \orgaddress{\street{} \city{Erlangen}, %\postcode{}, 
\country{Germany}}}

\abstract{
We introduce Temporal Fusion Nexus~(TFN), a multi-modal and task-agnostic embedding model to integrate irregular time series and unstructured clinical narratives. We analysed TFN in post-kidney transplant~(KTx) care, with a retrospective cohort of 3382 patients, on three key outcomes: graft loss, graft rejection, and mortality. Compared to state-of-the-art model in post KTx care, TFN achieved higher performance for graft loss~(AUC 0.96 vs. 0.94) and graft rejection~(AUC 0.84 vs. 0.74). In mortality prediction, TFN yielded an AUC of 0.86. TFN outperformed unimodal baselines~($\approx$10\% AUC improvement over time series only baseline, $\approx$5\% AUC improvement over time series with static patient data). Integrating clinical text improved performance across all tasks. Disentanglement metrics confirmed robust and interpretable latent factors in the embedding space, and SHAP-based attributions confirmed alignment with clinical reasoning. TFN has potential application in clinical tasks beyond KTx, where heterogeneous data sources, irregular longitudinal data, and rich narrative documentation are available.
}


\maketitle

\section{Introduction}\label{sec1}
%Background and problem
Kidney transplant~(KTx) recipients face risks of severe outcomes, including graft rejection episodes, graft loss, and mortality, despite advances in immunosuppression and monitoring~\cite{lemoine2019risk}. Early identification of patients at risk for severe post-transplant risks is vital to guide interventions and personalised follow-up care~\cite{schiffer2025smartntx, loupy2025advancing}. However, current clinical decision support tools for transplant prognosis are limited due to the complexity of post-transplant monitoring~\cite{wingfield2024clinical}. Electronic health records~(EHRs) encompass an array of structured data~\cite{survey2}~(e.g. vital signs, lab results, medication logs) and unstructured free-text~\cite{biobert}~(e.g. physician notes, discharge summaries, pathology and radiology reports) that are difficult to integrate for a combined analysis~\cite{sedlakova2023challenges}. Structured data alone provide an incomplete picture of patients and therefore clinicians rely heavily on narrative documentation of nuanced context, including clinical reasoning~\cite{seinen2022use}. The complementary nature of the modalities suggests that a multi-modal modelling approach can be beneficial. However, the irregular temporal structure of post KTx monitoring introduce challenges, including patient measurements collected at irregular time intervals, long-term temporal dynamics, and missing values~\cite{survey3}. Patterns of missing data can themselves be clinically informative~\cite{sim2025preserving}. For instance, a lab test might be omitted because it was deemed unnecessary, which itself conveys information about the patient's state. Moreover, inconsistencies and errors stemming from variations in clinical practices and manual data entry further complicate modelling efforts~\cite{garza2025error}. Additionally, model interpretability and explainability remain an open challenge for clinical explainability~\cite{xu2023interpretability}.

%Research gaps - using related works
Although there have recently been advances in deep learning approaches for EHR data, current approaches still struggle to capture long-term temporal dynamics, handle missing data without imputations, fully exploit complementary information from clinical narratives, and provide task-agnostic, interpretable representations~\cite{liu2023handling}. Deep learning has substantially improved outcome prediction from structured EHR data, including mortality~\cite{combining, combining2}, disease progression~\cite{behrt, dipole}, and hospital readmission~\cite{survey}. Longitudinally recorded features are modelled with RNNs~(e.g., LSTMs~\cite{ssi, fruitfly}) or their time-aware variants, i.e. T-LSTM~\cite{tlstm, tlstm_orig} and KIT-LSTM~\cite{kitlstm}, which handle irregular sampling. GRU variants with decay or attention also address missing data or sparsity~\cite{tan2020data, che2018recurrent}. However, RNNs frequently fail to retain long-term dependencies~\cite{shickel2019deepsofa}. Transformer-based models~(e.g., BEHRT, MedBERT~\cite{behrt, medbert}) improve representation learning through large-scale pre-training, but remain limited in modelling numerical values, complex longitudinal dependencies, and heterogeneous
dependencies across variables~\cite{zeng2022transformerseffectivetimeseries, bellamy2024labradorexploringlimitsmasked, zhang2023crossformer}. Self-supervised approaches including STraTS~\cite{tipirneni2022self} handle irregularly sampled clinical time-series data, however, modelling events as unordered triplets sacrifices explicit temporal ordering and weakens sensitivity to clinical progression. Hybrid approaches e.g.,  ATTAIN~\cite{attain} and Temporal Fusion Transformer~(TFT)~\cite{tft} address some limitations of RNNs and transformers by combining recurrence with attention, but their applicability to clinical trajectories remain constrained due to the inability to handle temporal irregularity. Transformers dominate in clinical NLP applications, including BERT-based models~(e.g.,BERT~\cite{bert}, ClinicalBERT~\cite{clinicalbert}, BioBERT~\cite{biobert}), and sentence transformers~(e.g.,GTE~\cite{gte}). Fine-tuned sentence transformers, e.g., Med-GTE-hybrid~\cite{medgte} achieve state-of-the-art performance on clinical narratives. Related works on multi-modal approaches that integrate text with structured EHR data have shown improvements in prediction across a range of clinical prediction tasks including length-of-stay, mortality, readmission, and surgery outcomes~\cite{combining, combining2, combining3, glaucoma, husmann, yang2022large}. However, in current multi-modal approaches, fusion of modalities is often late~(e.g., concatenating embeddings~\cite{ma2024global}), which leads to missing cross-modal dependencies. Attention-based fusion approaches are promising. Recent works that utilised attention-based fusion techniques~\cite{medmplm, fusingtransformer, temporalcrossattention} have proven that attention-based fusion facilitate alignment between modalities~(e.g., notes explaining lab changes~\cite{fusiontransformer2}), but the modelling remain task-specific, require large supervised labels, interpretability of models remains a challenge. Methods including attention visualisation~\cite{tft}, SHAP, and integrated gradients~\cite{survey2} offer partial insights into model interpretability, but do not address the underlying entanglement of latent factors. Disentanglement regularisation~\cite{disent_metrics, yao2024drfuse} offers promising direction by promoting latent structures that separate modality-specific and shared representations~\cite{tridira}, however, have rarely been applied to clinical multi-modal models and remain largely unexplored.

\begin{figure}[!htbp]
     \centering
     \includegraphics[width=\linewidth]{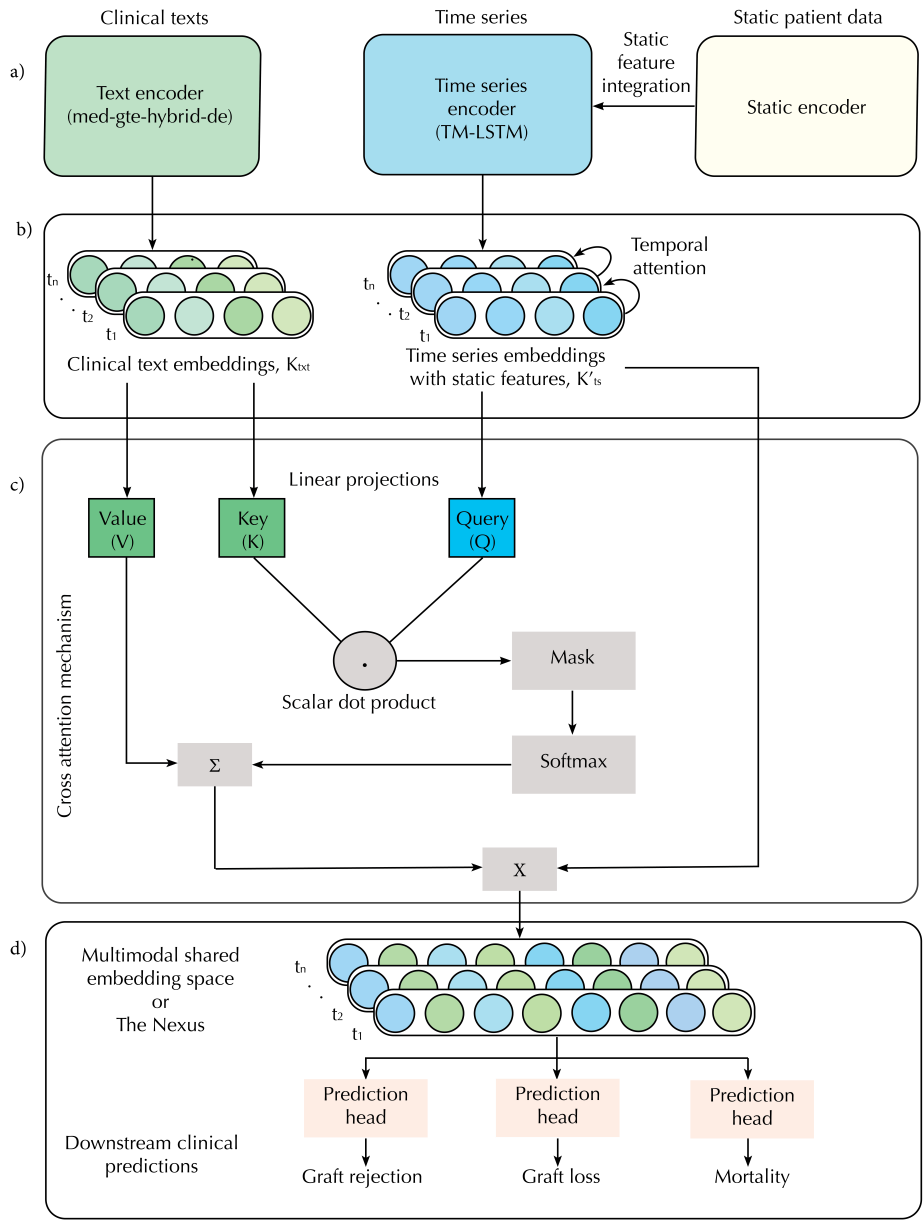}
     \caption{Overview of the TFN modelling approach. (a)~Modality specific modules, \textit{Med-GTE-hybrid-de} and \textit{TM-LSTM}, encode clinical texts and irregular time series data respectively. Static patient data is integrated with time series data through a variable-wise gating mechanism that selectively updates the \textit{TM-LSTM's} hidden state. (b)~Embeddings are generated for the text and time series modalities independently. The text embeddings are generated for each documented time point. The time series embeddings~(with integrated static data) are also generated for each documented time point. A multi-head self-attention mechanism with a causal mask~(temporal attention) is also utilised to capture the variable importance of different time points in patient's trajectory. (c)~Cross-attention mechanism is used to combine the clinical text embeddings and time series embeddings. Queries~(Q) are computed using the time series embeddings, while Keys~(K) and Values~(V) are obtained from linearly projected text embeddings. Interaction between the modalities are computed using scalar dot-product. (d)~The resulting representation~(attention output X) is a multi-modal shared embedding space, referred to as the Nexus. Downstream prediction performance is demonstrated on three key KTx outcomes: graft rejection, graft loss and mortality.}
     \label{overview}
 \end{figure}

%introduction of our approach
To address the aforementioned limitations, we introduce Temporal Fusion Nexus~(TFN), a multi-modal and task-agnostic modelling approach that integrates irregular time series data and clinical text into a disentangled embedding space. TFN comprises of two modality-specific modules: (i) a time-aware, sparsity-preserving LSTM with temporal attention, termed Temporal Masked LSTM (\textit{TM-LSTM}), designed to model irregular clinical time series with long-term temporal dependencies while retaining meaningful missingness patterns; and (ii) \textit{Med-GTE-hybrid-de}, a sentence-transformer fine-tuned on German clinical notes.   
In the \textit{TM-LSTM} module, we adopt an architecture inspired by T-LSTM~\cite{tlstm, tlstm_orig}, which uses memory decomposition with feature-wise decay, and GRU-D~\cite{che2018recurrent}, which introduces sparsity preserving feature-level mask for missing values. We incorporate both memory decomposition with feature-wise decay and sparsity preserving feature-level mask in the \textit{TM-LSTM} module.
The \textit{Med-GTE-hybrid-de} module is a gte-large~\cite{gte} based model, fine-tuned on German clinical text using an approach similar to Kumar et al.~\cite{medgte}. The med-gte-hybrid fine-tuning technique uses a combination of simple contrastive learning and denoising autoencoder, which outperforms other pretrained text models including clinicalBERT~\cite{clinicalbert} and longformer~\cite{beltagy2020longformer}. The two modules are integrated through a cross-attention layer~\cite{lu2019vilbert} with causal masking into a multi-modal shared embedding space, referred to as the Nexus. %The cross-attention fusion enabled the model to learn soft alignments between time-series events and relevant snippets of the text, effectively drawing insights from one modality to inform the other. 
Furthermore, we disentangle the Nexus to produce latent factors that are interpretable and attributable to specific data sources or clinical concepts. We achieve disentanglement in the Nexus by penalising correlated dimensions in the latent representations. We evaluate our TFN model on downstream predictive performance in three key KTx outcome predictions: graft loss, graft rejection, and mortality. Fig~\ref{overview} illustrates our modelling approach.\\

%contribution
The contributions of this work are as follows:
\begin{enumerate}
%modelling contribution
\item  We introduce the TFN model that integrates structured~(i.e., time-series and static patient data) via \textit{TM-LSTM}, and unstructured clinical data~(i.e., text) via \textit{Med-GTE-hybrid-de}. The TFN model handles irregular temporal data, long-term temporal dynamics, and missing data that might be clinically meaningful without imputation. We demonstrate through ablations that the TFN model learns informative temporal dynamics even under sparse sampling and that integration of modalities contributes meaningfully to predictive performance.
%evaluation of performance contribution
\item Our modelling approach is task-agnostic as the TFN model is trained entirely through self-supervised objectives, without relying on task-specific labels. When compared to the state-of-the-art, our TFN model achieves consistently higher predictive performance across three key clinical prediction tasks~(graft loss, graft rejection, mortality). 
%interpretability/explainability contribution
\item The TFN model learns disentangled latent representations that reduce redundancy and improve robustness, thereby improving interpretability. We demonstrate through disentanglement metrics and perturbation-based sensitivity analyses that the latent factors are more robust and less redundant than those of an entangled baseline. Additionally, we address model reliability through confidence calibration and explainability through SHAP-based feature attribution, which aligns closely with expert clinical reasoning. 

\end{enumerate}

\section{Results}\label{sec2}
\subsection{Cohort characteristics}
\begin{figure}[h]
     \centering
    \includegraphics[width=\linewidth]{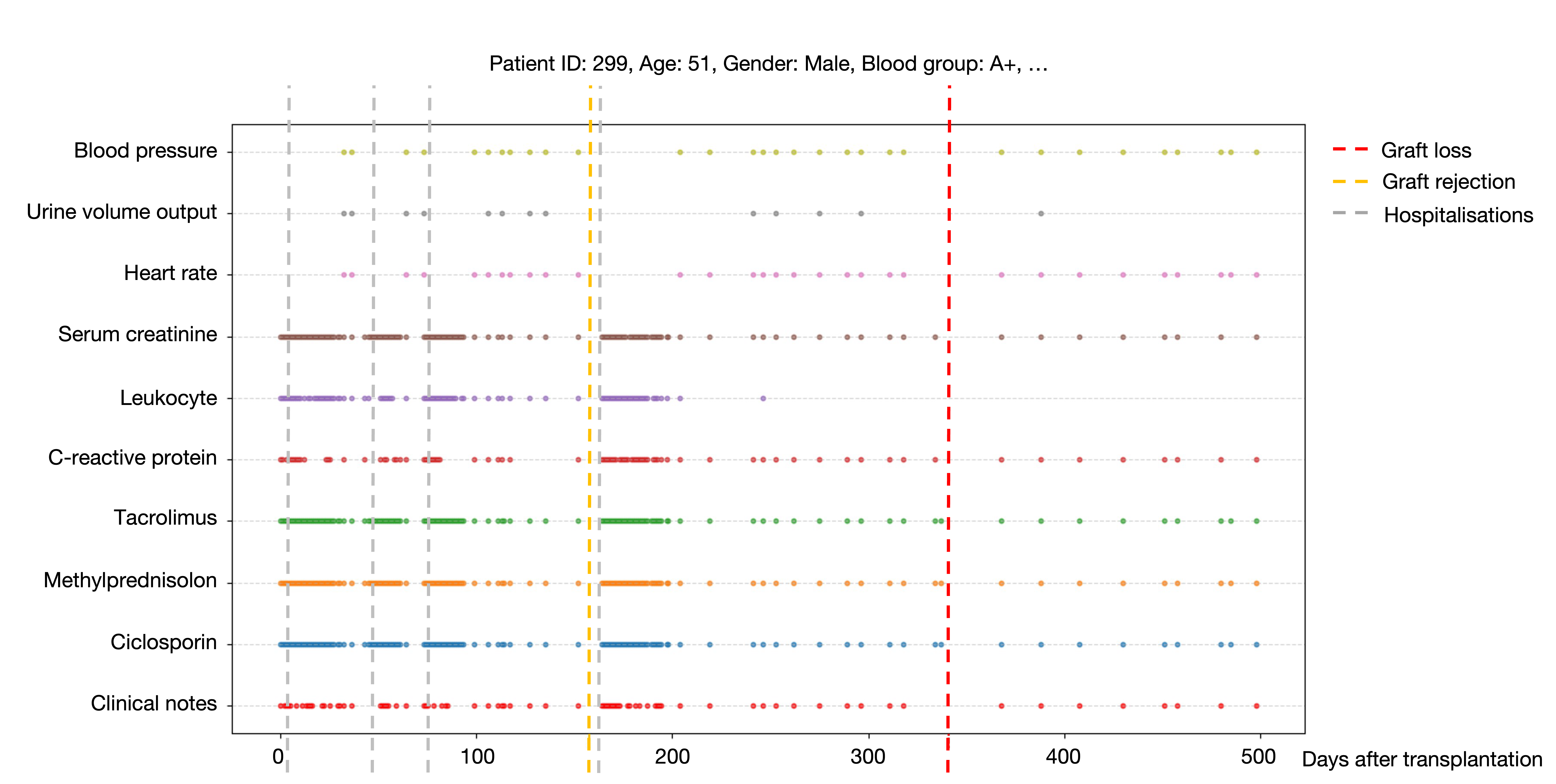}
     \caption{Trajectory of a representative KTx recipient in our cohort. Horiz. axis shows a relative time in days after transplantation. Static features included age, gender, blood group. Longitudinally collected time series features included blood pressure, urine volume, heart rate, serum creatinine, etc. Medications and clinical notes, which involves discharge summaries, radiology reports, etc. were documented longitudinally. Two key clinical outcomes are shown in the patient trajectory: graft rejection~(Day 163) and graft loss~(Day 341), besides several episodes of hospital admissions.}
     \label{patient_fig}
 \end{figure}

In this work, we used a subset of NephroCAGE dataset~\cite{nephrocage}, which comprises of 3382 kidney transplantations performed at Charité in Berlin between 2000 and 2020. Our cohort includes adult patients with a mean age of 54 years and a gender distribution of 62\% male and 38\% female KTx recipients.  The dataset contains longitudinal information, including laboratory measurements, medication records, and follow-up assessments, as well as recipient and donor demographics information. Fig~\ref{patient_fig} shows an example trajectory of a representative KTx recipient in our cohort with a selection of features. See Supplementary Tab.~1 for a complete list of features used.

\subsection{Prediction horizon for time series data}\label{2.2}
%short explaination of training for time-series
For the time series data, we used a self-supervised training scheme. To leverage the inherent temporal structure of patient data, we generated multiple training samples from each patient's longitudinal trajectory by systematically varying the temporal cutoff point along each patient’s longitudinal trajectory. We used the historical sequence up to the cutoff point as input to predict subsequent future time steps over a prediction horizon~(i.e., number of prediction steps during training). 
%~(i.e., sliding the cutoff along each patient’s longitudinal trajectory to generate multiple overlapping input–target pairs)
A decoder~(used only during training) reconstructed patient values following the cutoff time based on the encoded representation of all available historical data up to that point.

\begin{figure}[H]
    \centering
    \includegraphics[width=1\linewidth]{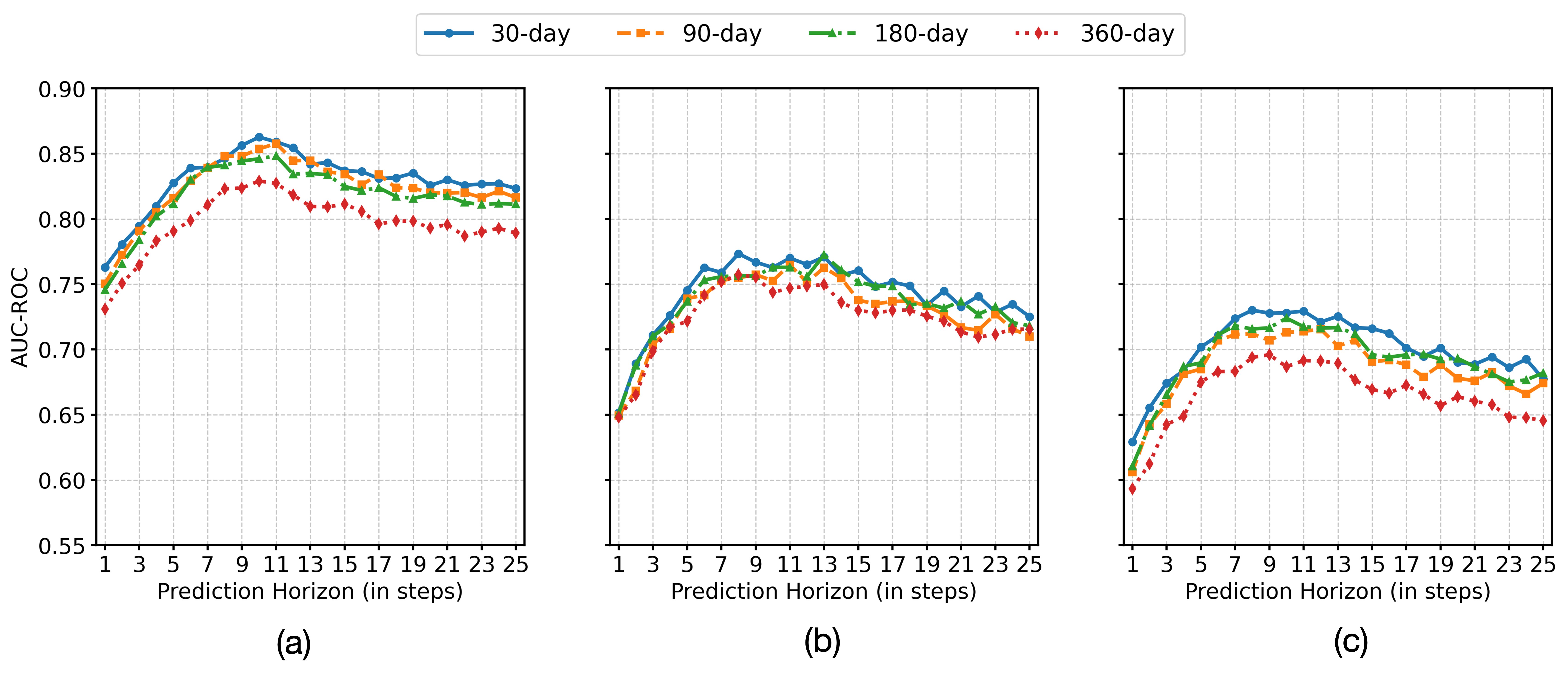}
    \caption{AUC of the multi-step prediction horizon during training of the time series encoder module \textit{TM-LSTM}. Three clinical prediction tasks were considered: (a)~Graft loss, (b)~Graft rejection, (c)~Mortality. Four prediction windows~(30 days, 90 days, 180 days, and 360 days) were analysed (see Sec.~\ref{2.2} for details). AUC for all three tasks peaked around 10 steps, which marks the empirically optimal context to capture clinically relevant temporal patterns.}
    \label{prediction_horizon}
\end{figure}

%how multi-step prediction horizon helps
Fig~\ref{prediction_horizon} shows the Area Under Curve~(AUC) of Receiver Operating Characteristic~(ROC) curve for downstream clinical prediction tasks~(i.e., graft failure, graft rejection and mortality) after training on the time-series data with different prediction horizons. The downstream clinical prediction heads were supervised and used the embeddings produced by the time series encoder module, \textit{TM-LSTM}. The prediction horizon in our case is defined as the number of future prediction steps during training, where one step correspond to the task-specific interval~(i.e., 30, 90, 180 or 360 days). For example, 1 step for the 30 day task means "predict 30 days into the future" and 1 step for 90 day task means "predict 90 days into the future". For the three clinical prediction tasks that we demonstrate, it can be observed that increasing the prediction horizon during model training initially improved performance, with a peak around 10 steps, before gradually declining thereafter. Therefore, we selected a prediction horizon of 10 steps for our final model training. Supplementary Fig. 1 demonstrates the effectiveness of TM-LSTM with prediction horizon of 10 steps in reconstruction of creatinine, which is a key biomarker for kidney function.

\subsection{Fine-tuning improves clinical text encoding}\label{2.3}
\begin{figure}[!htbp]
    \centering
    \includegraphics[width=1\linewidth]{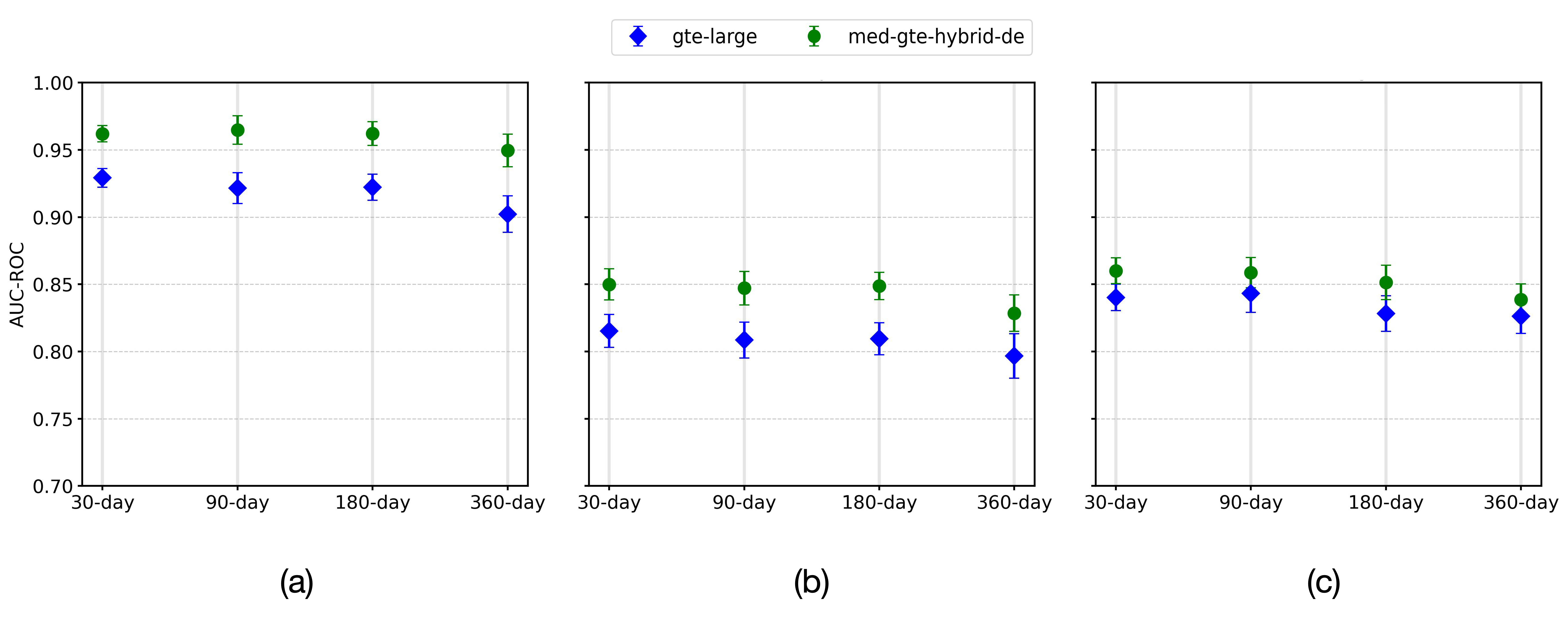}
    \caption{Performance comparison between our fine-tuned text encoder module \textit{Med-GTE-hybrid-de} and the base model, gte-large. AUC scores are plotted against four prediction windows~(30, 90, 180 and 360 days) across three prediction tasks: (a)~Graft loss (b)~Graft rejection (c)~Mortality. Error bars represent standard deviation across five-fold cross-validations. The \textit{Med-GTE-hybrid-de} model consistently outperformed GTE-large across all tasks and prediction windows, with the most notable improvement observed in graft loss prediction.}
    \label{finetuning_comparison}
\end{figure}

We compared the performance of our fine-tuned text encoder module, \textit{Med-GTE-hybrid-de} with the base model GTE-large on the three clinical prediction tasks. Here also, the downstream clinical prediction heads were supervised and used the embeddings generated by the two text encoders. Fig.~\ref{finetuning_comparison} shows the performance difference between the \textit{Med-GTE-hybrid-de} and the base gte-large model. For all three clinical prediction tasks and all four prediction windows~(i.e., 30 days, 90 days, 180 days and 360 days), the \textit{Med-GTE-hybrid-de} consistently outperformed the base gte-large model, with the most substantial improvements observed in graft loss prediction and the least in graft rejection prediction.

For graft loss prediction, \textit{Med-GTE-hybrid-de} achieved AUC scores approximately 0.03-0.05 higher than gte-large for all prediction windows. In graft rejection and mortality prediction tasks, the improvement was modest but still consistent, with gains of approximately 0.02-0.04 in AUC. The performance improvements demonstrate that domain-specific finetuning facilitated the text encoder, \textit{Med-GTE-hybrid-de}, to better capture clinically relevant features in German medical text.

\subsection{Integration of modalities shows performance improvements} \label{2.4}
\begin{figure}[H]
    \centering
    \includegraphics[width=1\linewidth]{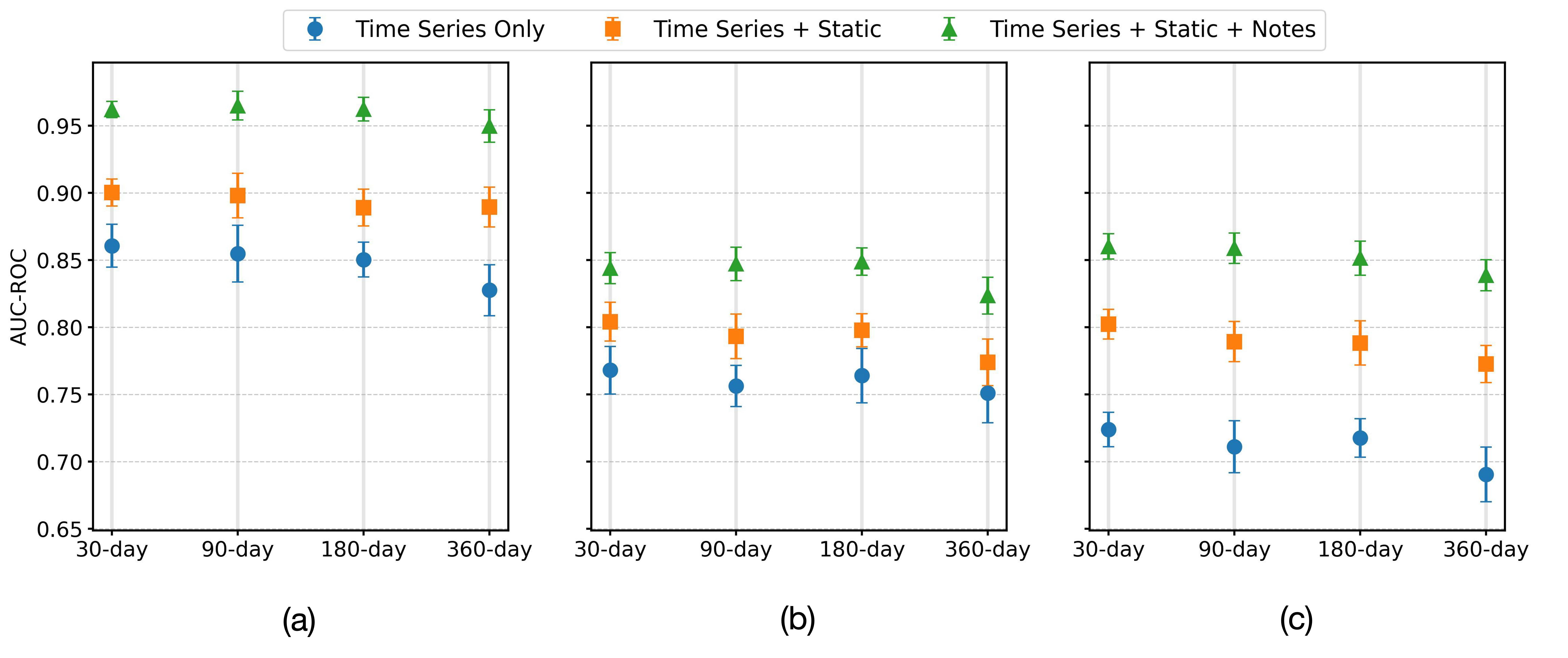}
    \caption{Illustration of additive effect of multi-modal data integration across three clinical prediction tasks: (a)~Graft loss (b)~Graft rejection (c)~Mortality. Four prediction windows~(30, 90, 180, 360 days) were analysed. We compare AUC of TFN model, which uses all three data modalities~(i.e., time series, static patient data and clinical notes) to a time series only baseline, and a time series with static patient data baseline. Error bars represent standard deviation across five-fold cross-validations. The consistent improvements in performance for TFN with all modalities demonstrate the additive value of each modality, and confirms that the TFN model could harness complementary information from the different modalities.}
    \label{data_sources_comparison}
\end{figure}

We compared the performance of the TFN model, which uses all three data modalities i.e., time series, static patient data, and clinical notes to: (a)~a time-series only baseline, and, (b)~a time-series with static patient data baseline, across the three downstream clinical prediction tasks. Fig.~\ref{data_sources_comparison} shows the performance metrics for the three downstream clinical prediction tasks. Again here, the downstream classifiers used the embeddings generated by the models. We report AUC scores for the three model variants. The performance of TFN remains robust across all prediction windows~(30, 90, 180, 360 days), with a slight but consistent decrease in the 360 days prediction window for all three tasks. The decline in 360 days prediction window can be attributed to the challenge of predicting clinical events over long time horizons, where intervening factors including new treatments or changes in patient status introduce additional variability.

The time series data alone provides a baseline performance across all tasks. For graft loss and rejection prediction, incorporating static data improves the AUC by approximately 0.03, while the further addition of clinical notes provides an additional improvement of approximately 0.06. In mortality prediction, the impact of static data is substantially larger, increasing AUC by approximately 0.1, while clinical notes contribute an additional improvement of 0.05. The different impact can be empirically explained through feature importance analysis~(see Sec.~\ref{ea_res}), which reveals that demographic features, mainly age, have greater predictive relevance for mortality outcomes compared to graft-related endpoints.

\subsection{Comparison with state-of-the-art} \label{2.5}
For the specific case of 90-day graft loss and rejection prediction, we benchmarked our TFN model against the Clinical Decision Support System~(CDSS) developed by Roller et al.~\cite{dfki}, which is based on the data from same source as NephroCAGE~(i.e., TBase system from Charité Berlin~\cite{tbase}), however the patient cohort is different. Notably, the CDSS from Roller et al. uses a random forest classifier with approximately 300 features requiring extensive pre-processing, whereas our model utilises only 26 features. 
Our TFN model outperformed the CDSS across all evaluation metrics, as detailed in Table~\ref{tab:graft-prediction}. For graft loss prediction, where the CDSS already achieved a high AUC of 0.94, our model demonstrate a modest improvement to 0.96. The improvement was substantial for graft rejection prediction, where our model achieved an AUC of 0.84 compared to 0.74. Our model showed superior sensitivity~(0.69 vs. 0.56) and specificity~(0.86 vs. 0.69) for graft rejection prediction too.

\begin{table}[htbp]
  \centering
    \caption{Performance comparison between our TFN model and Roller et al's CDSS~\cite{dfki} for 90-Day graft loss and rejection prediction tasks. TFN consistently outperforms Roller et al's CDSS across all metrics~(AUC,  sensitivity and specificity) for both prediction tasks. The performance gap is particularly pronounced for graft rejection prediction, where TFN achieves substantial improvements in AUC~(+0.1), sensitivity~(+0.13), and specificity~(+0.17). The results demonstrate that TFN was able to capture clinically relevant patterns for transplant outcomes with only 26 features without any explicit feature engineering compared to Roller et al's CDSS that uses more than 300 features with extensive feature engineering.}
  \label{tab:graft-prediction}
  \label{tab:graft-prediction}
  \begin{tabular}{@{}l @{\hspace{8mm}} l @{\hspace{10mm}} c @{\hspace{10mm}}c @{\hspace{10mm}} c@{}}
    \toprule
    \textbf{Task} & \textbf{Model} & \textbf{AUC} & \textbf{Sensitivity} & \textbf{Specificity} \\
    \midrule
    \multirow{2}{*}{90-day graft loss} 
      & Roller et al.~\cite{dfki} & 0.94 & 0.66 & 0.92 \\
      & TFN (this work) & \textbf{0.96} & \textbf{0.75} & \textbf{0.95} \\
    \midrule
    \multirow{2}{*}{90-day graft rejection} 
      & Roller et al.~\cite{dfki} & 0.74 & 0.56 & 0.69 \\
      & TFN (this work) & \textbf{0.84} & \textbf{0.69} & \textbf{0.86} \\
    \bottomrule
  \end{tabular}
\end{table}

\subsection{Model calibration}\label{2.6}
\begin{figure}[H]
\centering
\begin{subfigure}[b]{0.48\textwidth}
\centering
\includegraphics[width=\textwidth]{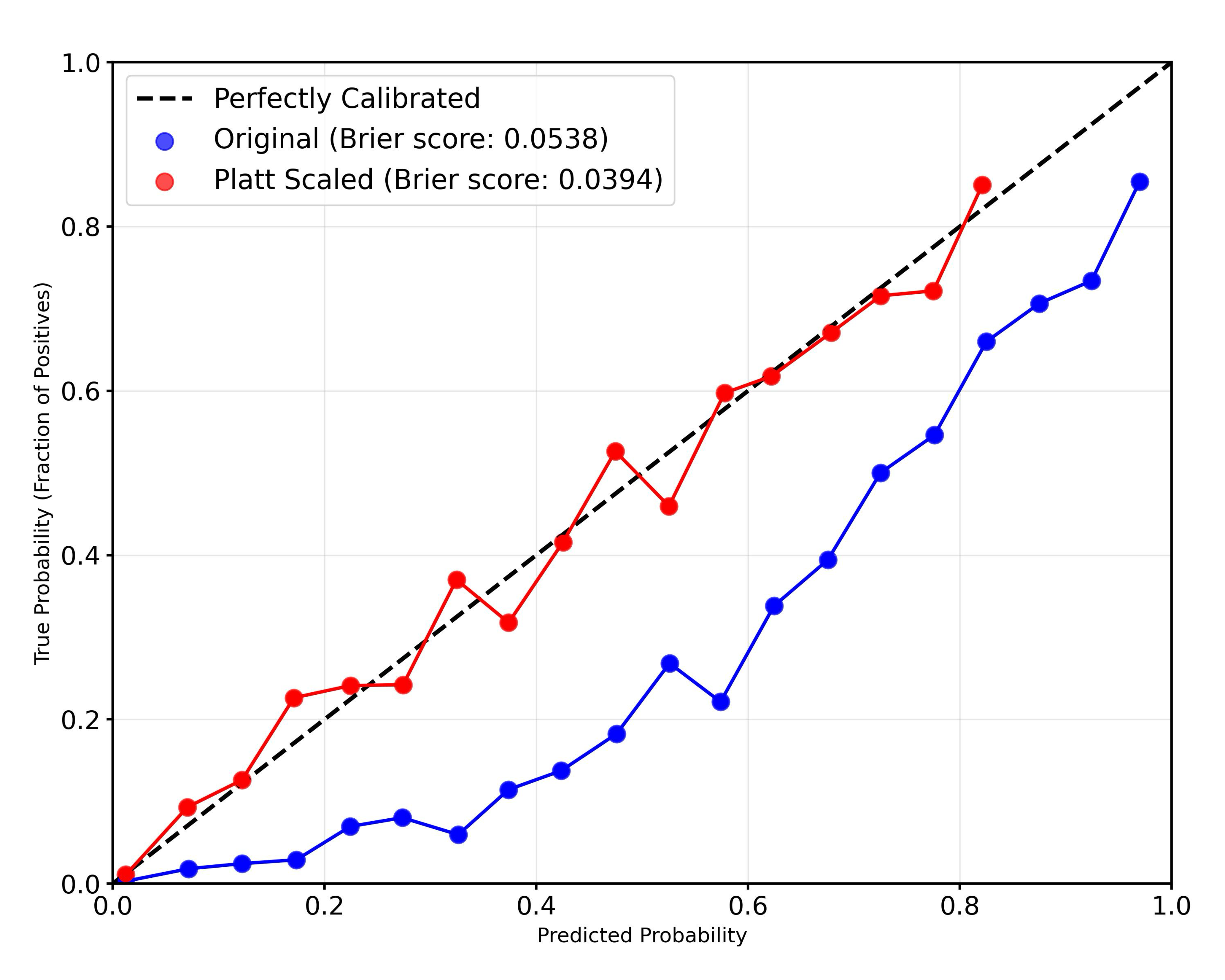}
\caption{}
\label{fig}
\end{subfigure}
\hfill
\begin{subfigure}[b]{0.48\textwidth}
\centering
\includegraphics[width=\textwidth]{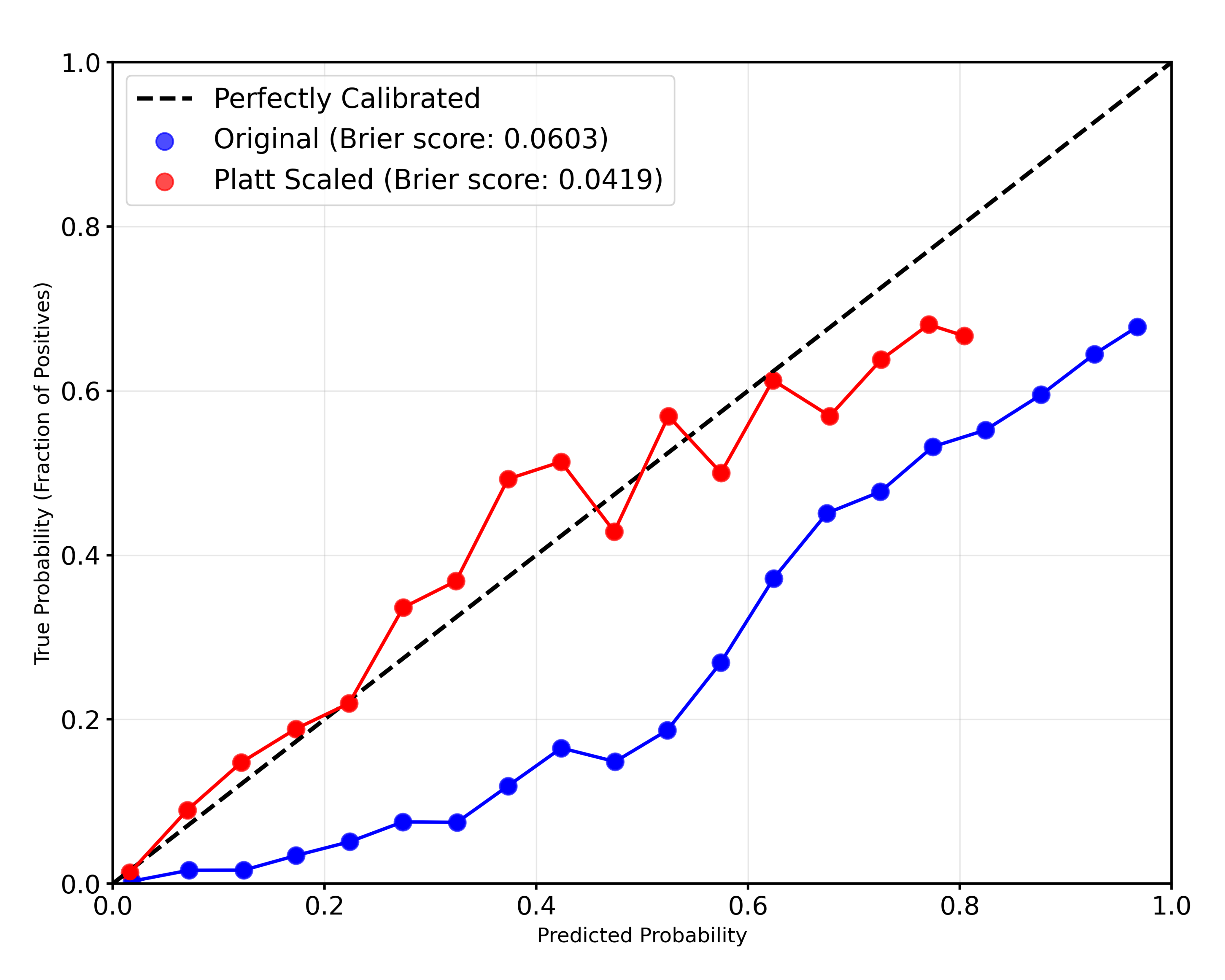}
\caption{}
\label{fig}
\end{subfigure}
\caption{Calibration curves showing the relationship between predicted probabilities~(horiz. axis) and observed frequencies~(vert. axis) for 20 equal-width probability bins. (a)~90 day graft loss. TFN outputs were initially overconfident and improved after Platt Scaling. (b)~90 day graft rejection Here too, TFN outputs were  initially overconfident and improved after Platt Scaling. When comparing calibration curves of (a) and (b), overconfidence was more pronounced for graft rejection. Brier scores reflect the mean squared difference between predictions and outcomes, with lower scores indicating better calibration.}
\label{calibration}
\end{figure}

The reliability of a data-driven model's probabilities is as important as the discriminative ability, especially in a clinical setting. However, confidence calibration is not directly captured by the discrimination metrics including AUC~\cite{sens}. Data-driven models tend to be poorly calibrated, often producing overconfident probability estimates that do not match empirical outcomes~\cite{calibration}. Therefore, we investigated the calibration of our TFN model using calibration curves. The deviation from the perfect calibration line~(dashed diagonal) was quantified by Brier scores of 0.0538 and 0.0603 for graft loss and rejection, respectively. Fig.~\ref{calibration} illustrates the calibration curves for 90-day graft loss~(Fig.~\ref{calibration}(a)) and rejection prediction~(Fig.~\ref{calibration}(b)). The curves plot the alignment between predicted probabilities and observed event frequencies. The blue curves represent the original TFN outputs, while the red curves demonstrate the effect of applying Platt scaling on TFN outputs.

For both prediction tasks, the original outputs of our TFN model exhibited overconfidence, particularly in the mid-to-high probability ranges~(0.4-1.0). For instance, when the TFN model assigned a probability of 0.8 for graft loss, only  $\approx$70\% of patients actually experienced the event. The overconfidence was particularly pronounced for graft rejection predictions, where an assigned probability of 0.8 correspond to an observed probability of 
$\approx$0.6. 

%Platt scaling improved calibration. The technique fits a logistic function to map the original predictions to better-calibrated probabilities without affecting the model's discriminative performance (i.e., AUC remains unchanged). 
When applying Platt scaling, the calibration improved and the TFN outputs more closely followed the ideal diagonal~(see calibration diagrams in Fig.~\ref{calibration}). Brier scores reduced correspondingly by approx. 27\% to 0.039 for graft loss and 30\% to 0.042 for graft rejection.

%The data points in the calibration plots represent 20 equal-width probability bins~(intervals of 0.05) along the predicted probability spectrum. Each point corresponds to the mean predicted probability within that bin~(x-coordinate) and the actual fraction of positive outcomes in that bin~(y-coordinate).
%Platt scaling reduces the model's overconfidence by moderating extreme probability predictions, particularly in the higher ranges. Due to the probability compression and the class imbalance in our dataset~(with relatively few positive cases), we didn't observe any of the data points in the highest probability intervals~($>$0.9) of the Platt-scaled curve, which reflected that the calibration process appropriately adjusted the model's excessive confidence given the underlying data distribution.

%Overall, the calibration analysis demonstrated that our model's original probability outputs required adjustment for optimal clinical utility, despite a good initial discrimination~(high AUC).

\subsection{Focus on clinically relevant temporal patterns} \label{2.7}
To assess the TFN's ability to capture temporal patterns in patient trajectory, we conducted an ablation study by introducing a temporal disruption~(i.e., randomly shuffling data samples along the timeline) during model evaluation. Table~\ref{tab:shuffle-comparison} presents the impact of the temporal discontinuity on model performance. We observed that under temporal discontinuity reconstruction error triples from 0.31 to 0.96, while AUC for 90-day graft loss and rejection prediction decreases by 0.25 and 0.22, respectively. The substantial performance degradation under temporal discontinuity confirms that TFN effectively leveraged sequential patterns of patient trajectories.

\begin{table}[htbp]
  \centering
  \caption{Analysis of temporal discontinuity on performance of TFN model. Arbitrary temporal discontinuity~(i.e., randomly shuffling of data sample timing) degraded both reconstruction capability~(average MSE of 0.96 vs. 0.31) as well as downstream task evaluation performance~(0.71 vs. 0.96 AUC for 90 day graft loss and 0.63 vs. 0.85 for 90 day graft rejection). The result demonstrates that TFN effectively utilises temporal dependencies of patient trajectories.}
  \label{tab:shuffle-comparison}
  \begin{tabular}{@{}lccc@{}}
    \toprule
    \textbf{Model} &
    \makecell{\textbf{Reconstruction error}\\\textbf{(Average MSE)}} &
    \makecell{\textbf{90-day graft loss}\\\textbf{(AUC)}} &
    \makecell{\textbf{90-day rejection}\\\textbf{(AUC)}} \\
    \midrule
    TFN        & \textbf{0.31} & \textbf{0.96} & \textbf{0.85} \\
    TFN (shuffled data timing) & 0.96          & 0.71          & 0.63 \\
    \bottomrule
  \end{tabular}
\end{table}

\begin{figure}[H]
    \centering
    \includegraphics[width=1\linewidth]{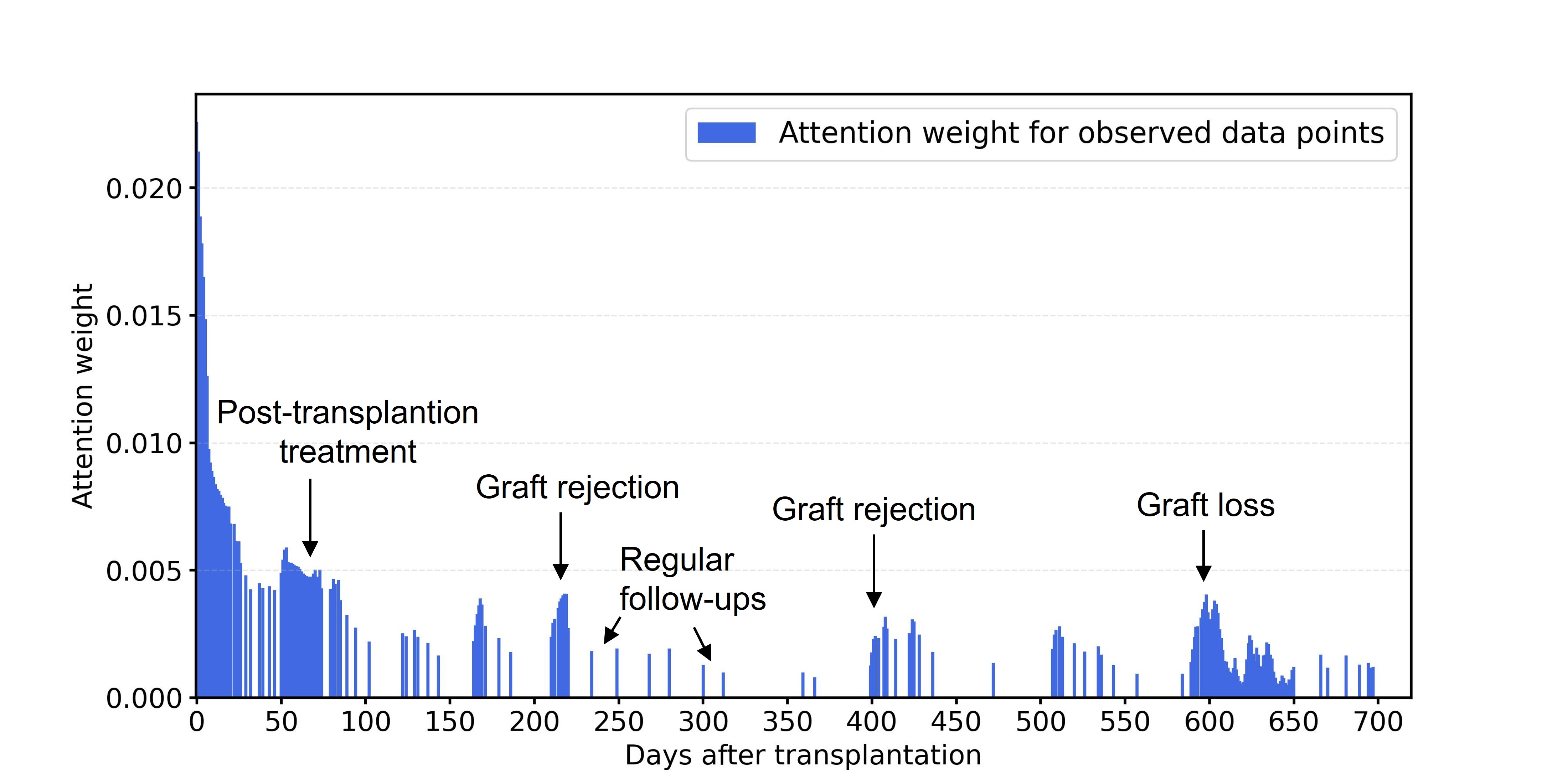}
    \caption{Illustration of temporal attention weight distribution for a representative KTx patient case. The attention mechanism assigned highest weights to clinically meaningful events: the early post-transplant period, days coinciding with documented hospitalisation periods~(Days 50 and 510), graft rejections~(around Days 210 and 400), follow-up treatment~(around Days 220-320), and graft loss~(around Day 600).}
    \label{Temp_attention_weights}
\end{figure}

Fig.~\ref{Temp_attention_weights} shows the temporal attention weights for a representative KTx patient case. The temporal attention weight distribution illustrates that the TFN model learned to focus on clinically meaningful temporal events, with high weights assigned to the post-transplant period~(days 0-30), followed by distinct peaks coinciding with graft rejection events and following treatment~(around days 220 and 400) and eventual graft loss~(around day 580). The attention pattern reveals that the TFN model learns to focus on relevant clinical events when computing patient representations.

\subsection{Disentanglement} \label{2.8}
Disentanglement i.e., encouraging latent dimensions to capture independent underlying factors, theoretically yield structured, interpretable, and meaningful representations~\cite{wang2024disentangled}. Therefore, we performed a disentanglement analysis to investigate if the Nexus of our TFN model can be made structured, interpretable, and clinically meaningful. Table~\ref{tab:dci} presents disentanglement analysis results using the Disentanglement, Completeness, and Informativeness~(DCI) framework~\cite{dci}. The baseline model with only reconstruction loss~($\mathcal{L}_{recon}$) achieves moderate disentanglement~(0.28) and completeness~(0.42) with adequate informativeness~(0.78). Addition of decorrelation loss~($\mathcal{L}_{decorr}$) improves disentanglement~(0.61) and completeness~(0.76), which suggest reduced redundant encoding. The final model with addition of disentanglement loss~($\mathcal{L}_{disent}$) further optimises disentanglement~(0.87) while maintaining informativeness~(0.85). 

\begin{table}[htbp]
  \centering
  \caption{Comparison of Disentanglement, Completeness and Informativeness metrics for three model variants: baseline model with only reconstruction loss, model with reconstruction and decorrelation loss, and model with reconstruction, decorrelation and disentanglement loss. It can be observed that adding $\mathcal{L}_{decorr}$ and $\mathcal{L}_{disent}$ to training leads to less redundancy~(higher disentanglement) and more robustness~(higher completeness). Informativeness values only slightly increase indicating that predictive power is conserved.}
  \label{tab:dci}
  \begin{tabular}{@{}lccc@{}}
    \toprule
    \textbf{Model} &
    \makecell{\textbf{Disentanglement}}&
    \makecell{\textbf{Completeness}} &
    \makecell{\textbf{Informativeness}} \\
    \midrule
    $\mathcal{L}_{recon}$ & 0.28 & 0.42 & 0.78 \\
+ $\mathcal{L}_{decorr}$ & 0.61 & 0.76 & 0.83 \\
+ $\mathcal{L}_{disent}$ & 0.87 & 0.78 & 0.85  \\
    \bottomrule
  \end{tabular}
\end{table}

The increase in informativeness across model variants~(0.78 vs. 0.85) indicates that the disentanglement process preserved predictive power while re-organising the structure of the Nexus. Higher disentanglement values correspond to  concentrated feature influence in smaller latent regions, which enhances model robustness.

Further support for the effectiveness of our disentanglement approach comes from perturbation sensitivity analysis~(see Supplementary Table 2), where we systematically introduced noise to individual clinical features and observed the impact on different dimensions of the latent representation. The complete model with disentanglement loss showed localised sensitivity patterns, with perturbations to specific features affecting a small subset of latent dimensions in the Nexus. In contrast, the baseline model showed diffused sensitivity patterns, where perturbation of a single feature produced widespread changes across numerous latent dimensions in the Nexus.

\begin{figure}[H]
    \centering
    \begin{subfigure}[b]{0.47\textwidth}
        \centering
        \includegraphics[width=\textwidth]{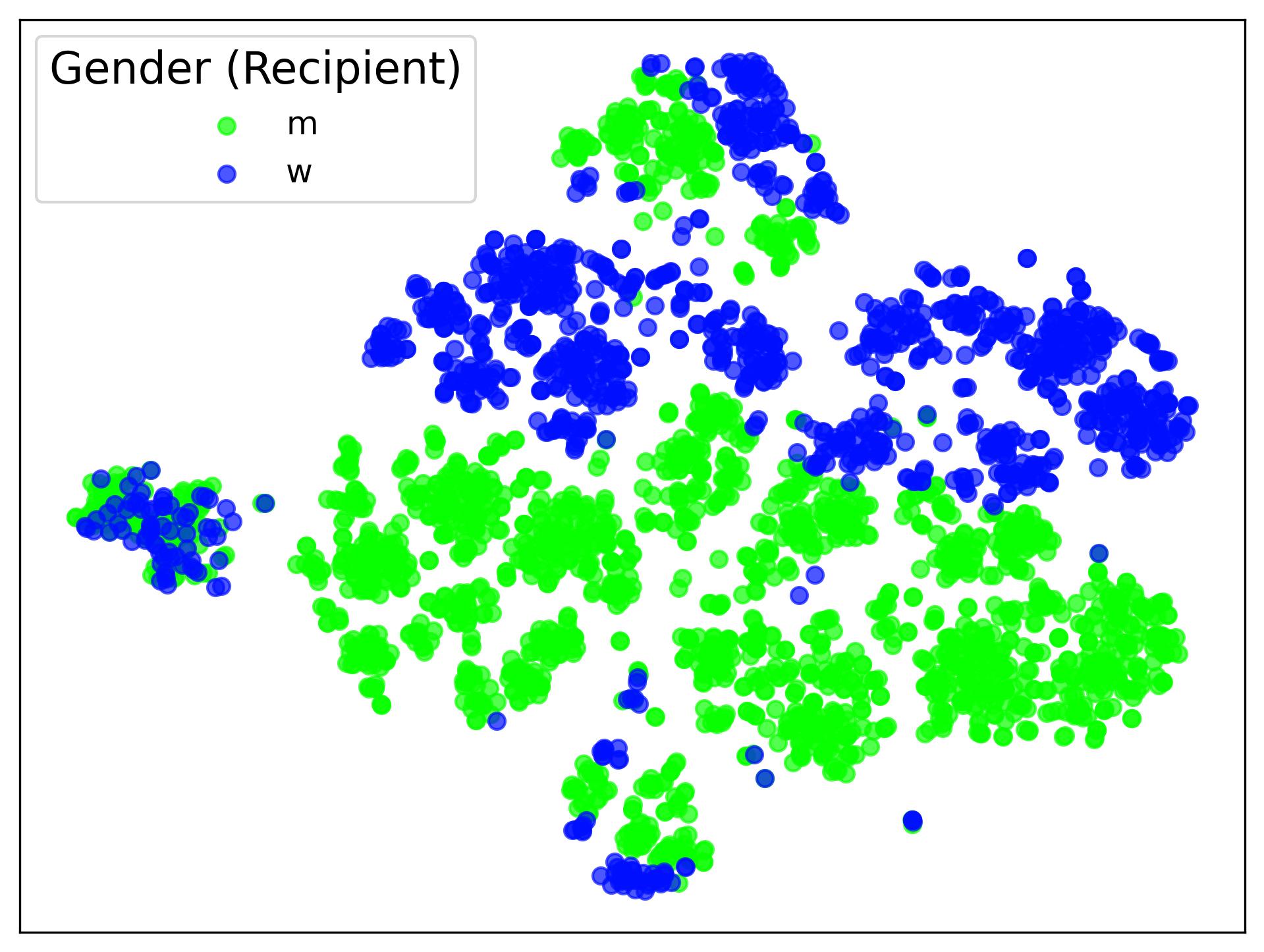}
        \caption{}
        \label{fig:top-left}
    \end{subfigure}
    \hfill
    \begin{subfigure}[b]{0.47\textwidth}
        \centering
        \includegraphics[width=\textwidth]{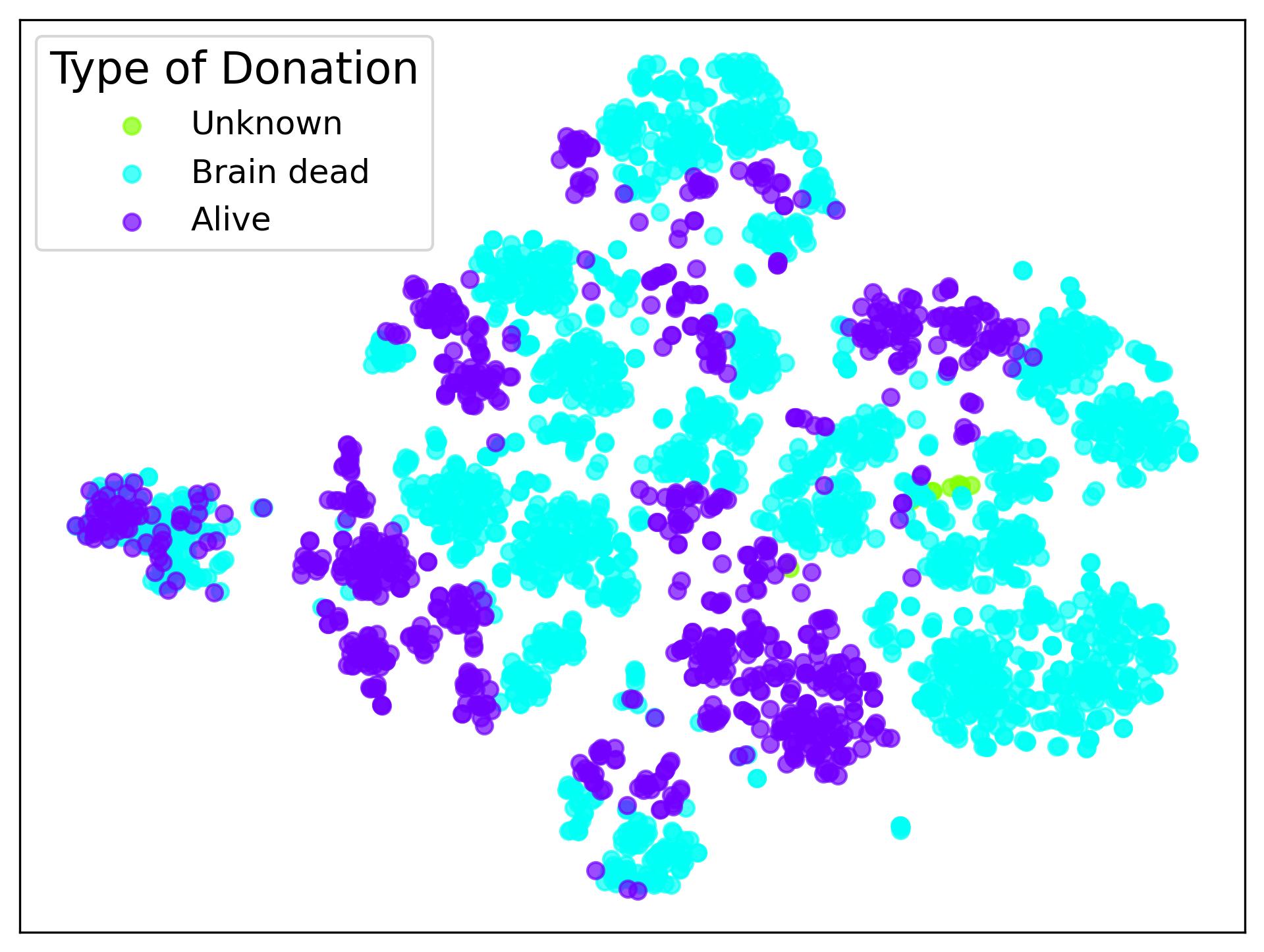}
        \caption{}
        \label{fig:top-right}
    \end{subfigure}
    \vskip\baselineskip
    \begin{subfigure}[b]{0.47\textwidth}
        \centering
        \includegraphics[width=\textwidth]{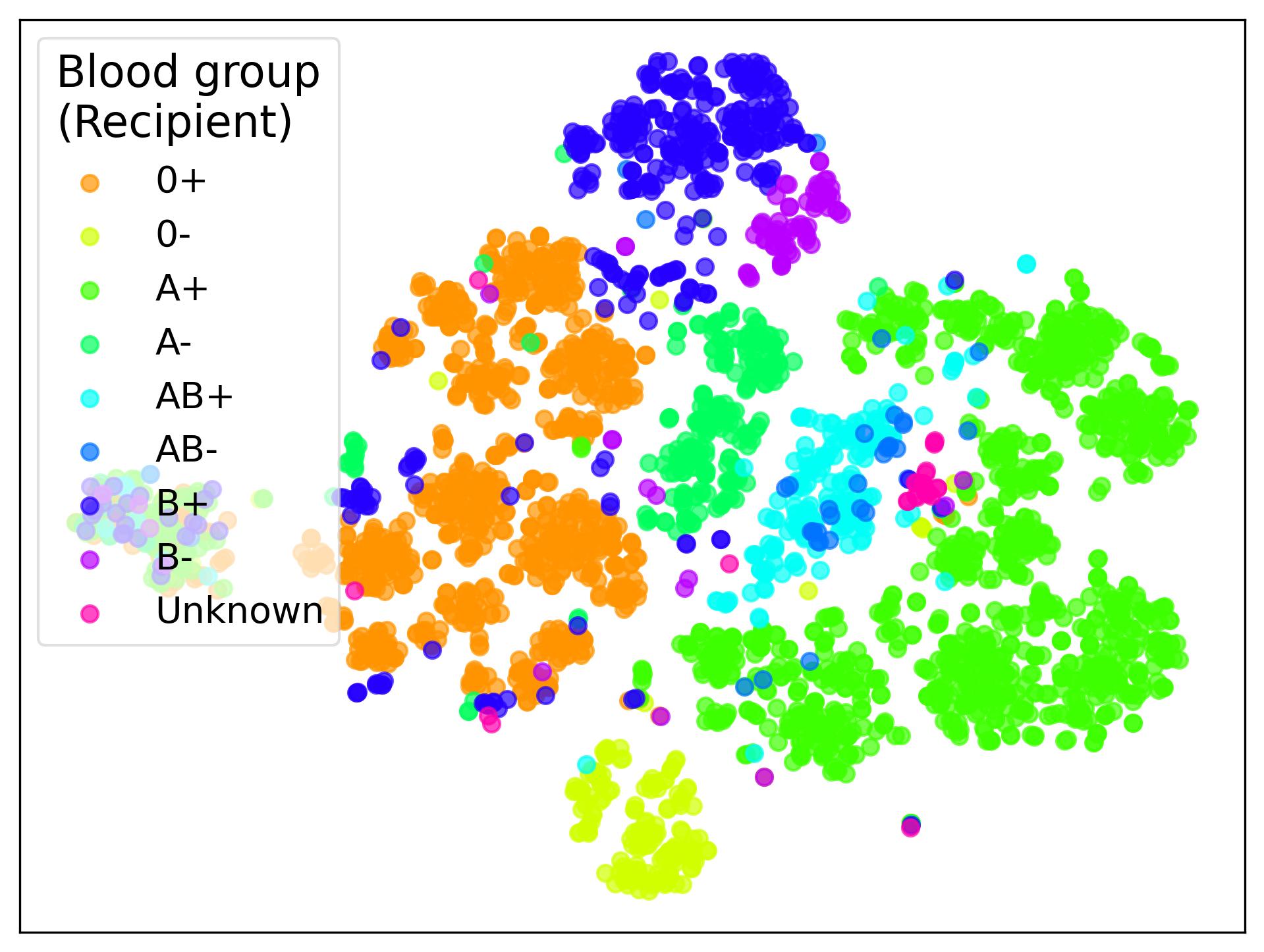}
        \caption{}
        \label{fig:bottom-left}
    \end{subfigure}
    \hfill
    \begin{subfigure}[b]{0.47\textwidth}
        \centering
        \includegraphics[width=\textwidth]{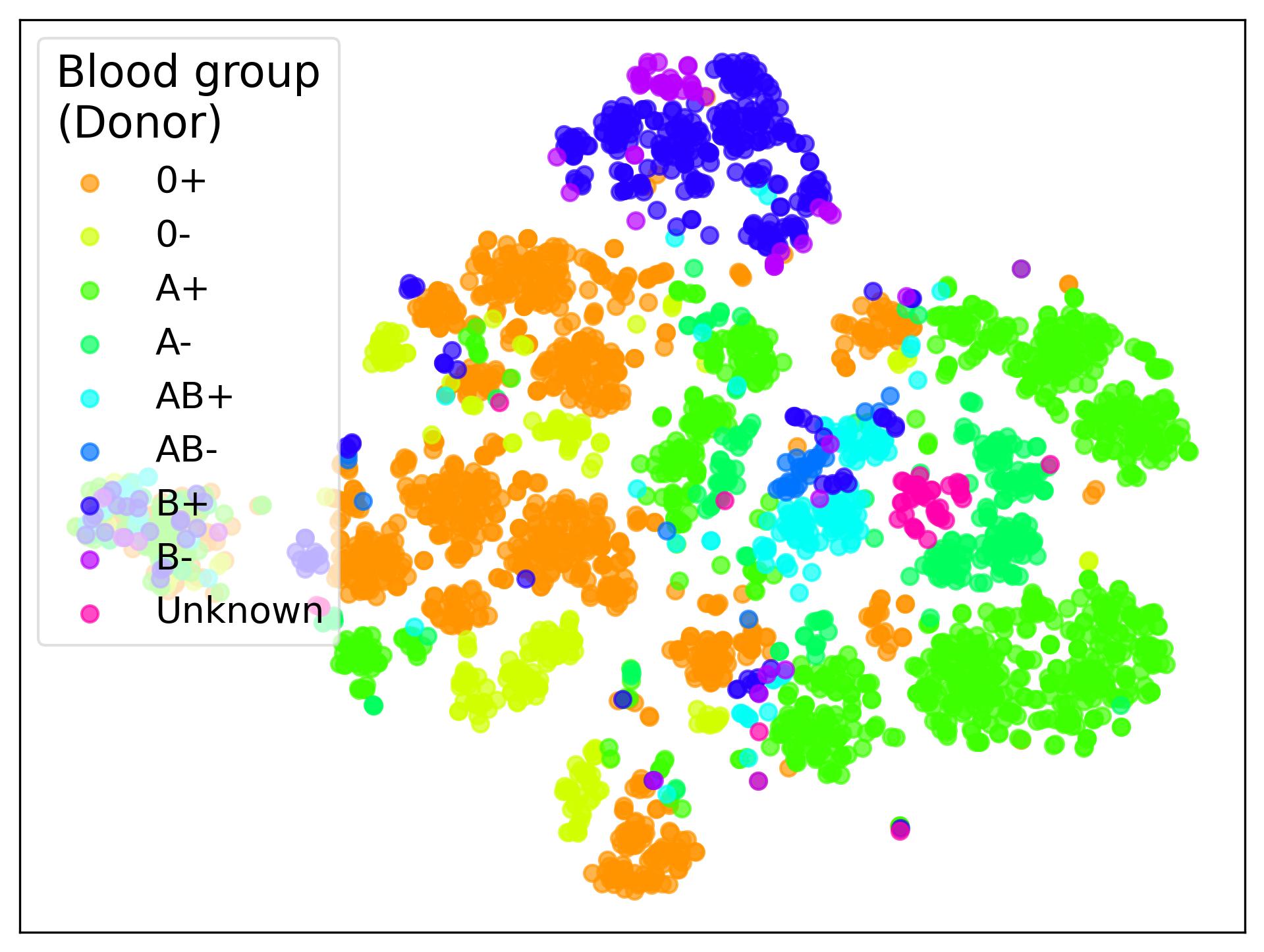}
        \caption{}
        \label{fig:bottom-right}
    \end{subfigure}
    \caption{t-SNE projection of the disentangled embedding space learnt by TFN. Each point represents a patient and is colour-coded by a distinct clinical feature category. The visualisations show that the latent space exhibits intrinsic structure without any explicit clustering. (a)~Separation emerging along KTx recipient's gender. (b)~Separation emerging by donor type~(living vs. deceased). (c)~Separation emerging by KTx recipients's blood group. (d)~Separation emerging by donor's blood group.}
    \label{tsne}
\end{figure}

Fig.~\ref{tsne} visualises our TFN model's disentangled embedding space through t-SNE. The t-SNE analysis shows distinct clusters by gender, donation type, and blood groups. The visualisation reveals an intrinsic structure in the learnt TFN representations that align with clinically relevant patterns.

\subsection{Explainability analysis}\label{ea_res}
\begin{figure}[!htbp]
\centering
\includegraphics[width=0.85\linewidth]{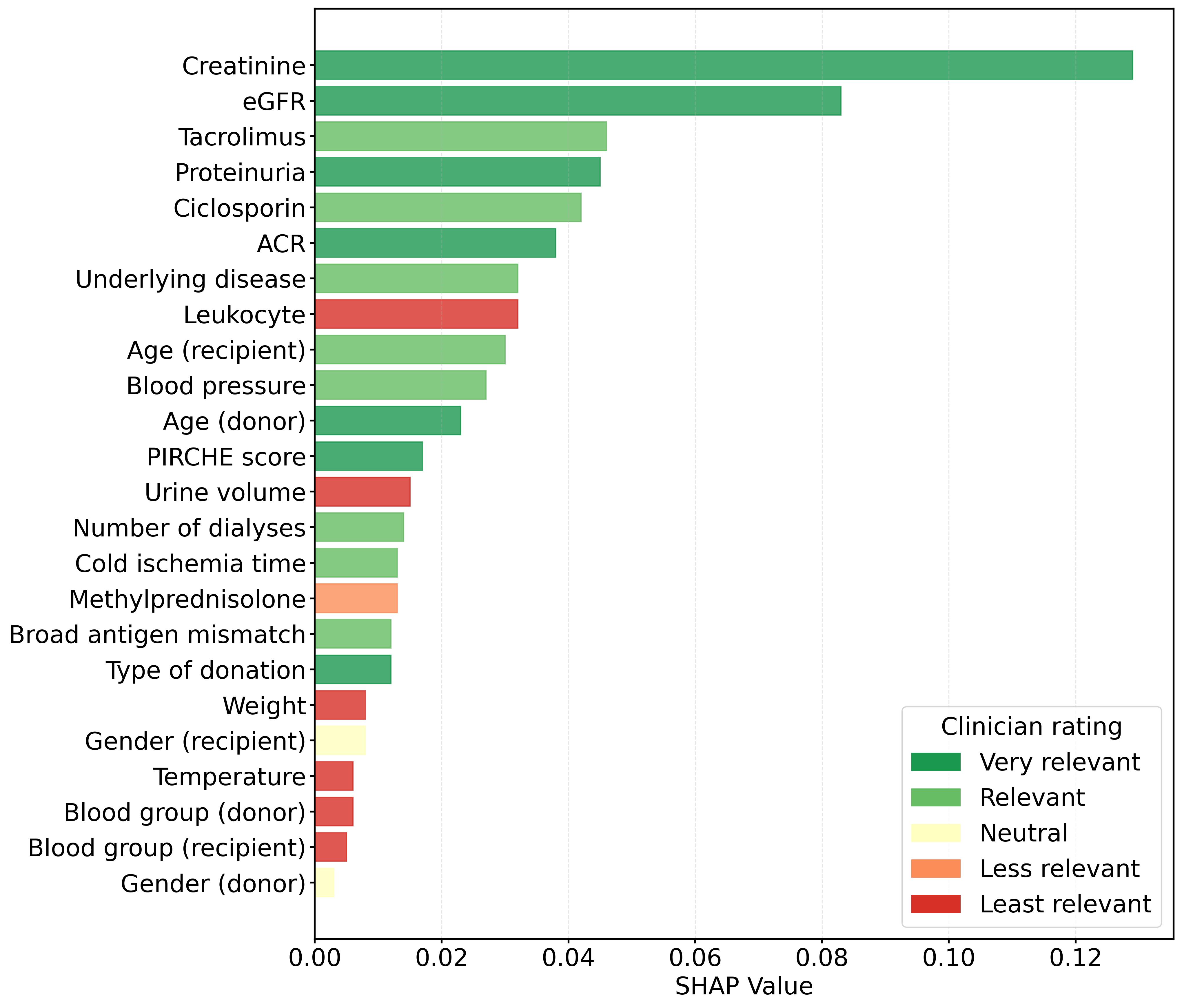}
\caption{Mean absolute SHAP values for each feature, indicating their relative importance in TFN predictions for graft loss. The relevance rating of a clinical expert of each feature is shown too. Kidney function markers~(creatinine and eGFR) appear as top features in SHAP-based feature importance, followed by immunosuppressant levels and indicators of kidney damage~(proteinuria, ACR). The moderate correlation~($\rho=0.62$) between SHAP values and clinician ratings demonstrate substantial alignment between the model's learnt feature importance and clinical expertise.}
\label{shap}
\end{figure}

Fig.~\ref{shap} presents the SHAP values along with the clinician ratings for the downstream graft loss prediction task using our TFN model. The clinician ratings were provided by a clinical expert with more than 20 years of experience, who independently assessed each input feature’s predictive relevance for the three downstream tasks using a 5-point Likert scale ranging from 1~(very relevant) to 5~(not relevant). 

Creatinine and eGFR were the most influential features for TFN, which aligns with established clinical knowledge and related work~\cite{dfki}. Creatinine and eGFR are direct indicators of kidney function, and deteriorating values are associated with impending graft failure. The model also assigns substantial importance to immunosuppressive medications~(Tacrolimus, Ciclosporin), proteinuria, and albumin-to-creatinine ratio~(ACR), all of which were rated as clinically relevant by our expert. The notable disagreement between model-derived importance and clinician ratings is for leukocyte count.

The alignment between model-derived importance and clinical expertise is notable, with features deemed most important by the model generally receiving high relevance ratings from the clinician. Similarly, features assigned lower importance by our model, including gender and blood group, were independently rated as less clinically relevant by our expert. The Spearman rank correlation coefficient between SHAP values and clinician ratings is $\rho=0.62$ for graft loss prediction, which indicate a moderate to high correlation between algorithmic and expert feature priority.
Similar patterns of alignment between the model and the clinical expert are also observed for graft rejection~($\rho=0.54$) and mortality prediction~($\rho=0.42$), but with less correlation strength. 
For the corresponding graphs, see Supplementary Fig. 2 and 3.. 

\section{Discussion}\label{sec3}
In this work, we developed TFN model, a task-agnostic multi-modal embedding model to integrate irregular time series and unstructured clinical texts from EHRs. The modelling approach utilised a temporal module, \textit{TM-LSTM}, and a text module, \textit{Med-GTE-hybrid-de}. The two modules were contextualised using temporal information and then integrated using a cross-attention mechanism. The training of TFN did not rely on any task-specific objectives, making it task-agnostic. \textit{TM-LSTM} module was trained in a self-supervised manner to predict future time steps, and \textit{Med-GTE-hybrid-de} module learnt through a combination of contrastive learning and denoising autoencoder. Thus, TFN learnt general purpose patient representations independent of any prediction task. 

We evaluated TFN in post KTx care scenario on a retrospective cohort of 3382 KTx recipients for three key outcomes~(graft loss, graft rejection, and mortality) for KTx recipients. 
%\subsection{Model performance}
%Our TFN model showed consistently relevant predictive performance across all downstream clinical prediction tasks. 
The downstream predictive performance of TFN was exceptional for graft loss, exceeding AUC of 0.95, and above 0.85 for graft rejection and mortality~(see Sec.~\ref{2.4}). Additionally, TFN outperformed Roller et al.~\cite{dfki}'s random forest based CDSS in 90-day graft loss and 90-day graft rejection prediction tasks~(see Sec.~\ref{2.5}) with higher AUC values~(0.96 vs. 0.94 for 90-day graft loss, and 0.84 vs. 0.74 for 90-day graft rejection) while using fewer input features~(26 vs. 300). The TFN performance suggests that its architecture could extract more information from fewer variables than the CDSS of Roller et al. by capturing complex clinical interactions and temporal patterns. Moreover, our modelling approach required less feature engineering. 
%Notably,  Roller et al.'s CDSS was trained on a different subset of data collected using TBase~\cite{tbase}.  A direct comparison on the same data subset was not feasible due to the unavailability of the engineered features used in the work of Roller et al. Because of the different evaluation datasets, TFN's performance advantage should be interpreted with caution. A more definitive comparison would require evaluating both modelling approaches on an identical dataset.

%\subsection{Effects of multi-modality in modelling process}
The multi-modality of TFN contributed to predictive performance~(see Sec.~\ref{2.4}). %We confirmed that complementary information from different data sources improved model accuracy. 
The time series only baseline was outperformed by the model using time series with static data, with AUC gains of approximately 0.03 for graft loss and rejection, and 0.1 for mortality. Incorporating clinical notes yielded further improvements, adding roughly 0.06 AUC for graft loss and rejection and 0.05 for mortality. Consistent gains demonstrated the additive value of each modality. Although we did not include a text only baseline in the main experiments, we assessed text only performance during fine-tuning of our text encoder, \textit{Med-GTE-hybrid-de}~(see Sec.~\ref{2.5}). The \textit{Med-GTE-hybrid-de} model achieved results comparable to TFN across all tasks~(AUC $\approx$ 0.95 for graft loss and $\approx$ 0.85 for graft rejection and mortality). However, \textit{Med-GTE-hybrid-de} exhibited performance degradation in the 360-day prediction window, with AUC values for graft rejection and mortality falling below 0.85, likely due to its limited ability to capture long-term temporal structure. The TFN model performed better than the \textit{Med-GTE-hybrid-de} in 360 prediction window across all three clinical prediction tasks. Therefore, it could be reasoned that TFN was able handle the long-term temporal dynamics better compared to text only baseline because of inclusion of time series modality, multi-step prediction horizon and the temporal attention mechanism at play in the temporal module, \textit{TM-LSTM}.

%\subsection{Temporal patterns}
Our TFN model was able to learn clinically meaningful temporal patterns, which we validated using an ablation study, where temporal disruption was introduced~(see Sec.~\ref{2.7}). A substantial drop in performance was observed when the temporal ordering was disrupted. We also analysed temporal attention weights of TFN, which revealed that the model focused on clinically relevant time points. The emergence of clinically relevant attention patterns without explicit specification suggested that TFN learnt meaningful representations of patient trajectories through self-supervised training. Furthermore, the analysis on prediction horizon during training of time series encoder~(i.e.,\textit{TM-LSTM}) showed that a multi-step prediction horizon~(i.e., model predicts multiple steps into the future) during training encouraged our model to learn relevant clinical patterns~(see Sec.~\ref{2.2}).  However, beyond the prediction horizon of 10 steps, the benefit saturated and eventually diminished, which suggested that the prediction horizon of 10 was a natural scope for capturing relevant clinical patterns in the dataset. %Another key observation is that the long-term prediction tasks~(i.e., outcomes in 360 days preditcion window) shows more relative improvement with increased prediction horizons, which indicate that models trained to capture extended temporal patterns are particularly valuable for forecasting distant clinical events.
Long-term prediction tasks~(e.g., 360-day outcome predictions) showed substantial improvement with increased prediction horizons, indicating that the \textit{TM-LSTM} model learnt to capture long-term temporal dynamics. The stagnation of performance after 10 steps can be attributed to the increasing difficulty of modelling long-range dependencies~($>$10) in the dataset. %It is worth highlighting that a prediction horizon of 10 was optimal in our experiments, although it may be dataset-dependent. 

%\subsection{Interpretability and explainability}
Disentanglement approach improved the structure of the Nexus, as quantified by DCI metrics~(see Sec.~\ref{2.8}). Perturbation sensitivity analysis confirmed the improved structure of the Nexus, demonstrating that feature effects localised to latent dimension subsets. Thus, disentanglement improved model robustness to input variations and facilitated interpretability. Additionally, the t-SNE visualisations of the Nexus revealed natural clustering according to clinically relevant categories. The emergent clustering occurred without explicit clustering objective, which confirmed that the model learnt a meaningful separation of patient characteristics within the Nexus. %The evaluation primarily examined how individual features related to each latent dimension. However, clinical variables exhibit complex interdependencies, which latent dimesnions capture as hidden representations. The hidden representations encode clinically meaningful patterns that reflect a patient's trajectory. Consequently, achieving a disentanglement aligned strictly with original input features was neither realistic nor desirable. Instead, the objective was to capture and localise relevant latent factors. Although the objective is inherently challenging because many clinically relevant patterns are multi-factorial and not directly observable from single input variables.

The integration of clinical notes through \textit{Med-GTE-hybrid-de} introduced interpretability challenges. Structured data features facilitated relatively straightforward attribution analysis, however, text embeddings lacked direct correspondence to specific semantic concepts. Thus, the contribution of particular textual elements to predictions remains opaque despite a demonstrable predictive value.

SHAP-based attributions demonstrated alignment between model-derived feature importance and clinical expertise, particularly for key kidney health indicators, creatinine and eGFR~(see Sec.~\ref{ea_res}). However, there were also disagreements between clinical expert's rating and SHAP-based feature importance, e.g., leukocyte count. 
%The notable disagreements are leukocyte counts and heart rate, which are considered important by our model, but rated "least relevant" by the clinical expert. The heart rate is particularly interesting because recent studies have shown that cardiovascular health, which also includes heart rate, is a major contribution factor for graft outcomes~\cite{weiner2012kidney}. 
Moreover, we observed weaker correlation between model and expert for mortality prediction compared to graft loss and graft rejection predictions. The weaker correlation observed for mortality prediction could have been be due to the multi-factorial nature of mortality outcome, where our dataset primarily captured kidney-specific metrics rather than broader health indicators. Furthermore, we only had ratings from a single clinical expert, while highly valuable, might have limited the generalisability of the assessments. Expanding the validation to include assessments from multiple clinicians could strengthen our findings and address the potential subjectivity. Future work could incorporate qualitative analyses of individual features where the model’s assessments diverge from clinical expert's assessments. Another promising direction would be to investigate global, model-level explainability, beyond local prediction-level attributions to provide a holistic understanding of how the model created representations of the data.

%\subsection{Model reliability}
Well-calibrated model probabilities provide clinicians with trustworthy risk assessments for transplant patients~\cite{reliable}. Neural networks typically produce poorly calibrated probabilities. We observed overconfidence in our TFN's probability estimates and applied Platt scaling to align predicted probabilities with observed frequencies~(see Sec.~\ref{2.6}). The initial miscalibration was most pronounced in mid-to-high probability ranges, likely due to class imbalance in the dataset. As our dataset contained few positive cases~(corresponding to graft loss, graft rejection and mortality), which made reliable pattern learning from positive cases challenging. Here, oversampling mitigated the class imbalance during training. %Future works on model reliability could explore uncertainty estimation or conformal prediction methods.

%\subsection{Clinical challenges}
Clinical reality introduces inter-relationships between prediction tasks despite conceptual independence. For example, in our analysis, the definition of graft loss excluded patients, who died with functioning grafts. However, mortality and graft dysfunction share common risk factors~\cite{merzkani2022death}. Patients experiencing graft rejection typically demonstrate higher probabilities of subsequent graft loss and mortality~\cite{levitsky2017acute}. Notably, our TFN model demonstrated better predictive performance for graft loss prediction compared to graft rejection prediction, which might have been because graft loss showed more consistent patterns in the dataset. Furthermore, our evaluation relied on data from a single transplant centre, potentially introducing institutional biases in clinical practices and documentation. Future work should explore a multi-centre setup, potentially utilising federated learning. The ongoing smartNTx clinical study~\cite{schiffer2025smartntx} is geared to gather multi-centre KTx data. 

A challenge in clinical data analysis is the inability to definitively determine whether missing data follows a missing at random~(MAR) or missing not at random~(MNAR) mechanism from observed data alone. Missingness patterns in medical data often encode clinically relevant information, for instance, reduced monitoring for stable patients. We therefore used a sparsity preserving feature-level masking, rather than imputation, to allow our TFN model to utilise missingness as an informative signal, while avoiding potential bias~(e.g., averaging effects introduced by imputation). %Alternative approaches could have included various imputation methods or complete case analysis. However, alternative approaches would either assume MAR conditions or substantially reduce the sample size.

%\subsection{Applicability to other clinical domains and outlook}
TFN modelling could be applied to other clinical domains~(e.g., oncology, cardiology, intensive care medicine) that similarly involve heterogeneous data sources, irregular longitudinal measurements, and rich narrative documentation. Future work should validate TFN across multiple institutions and healthcare systems to ensure generalisability. Differences in documentation practices, clinical workflows, data standards~(e.g., FHIR, openEHR), as well as variability in note structure and terminology may affect model performance and necessitate domain-specific adaptation. Moreover, data quality issues e.g., lack of unique patterns across clinical settings, inconsistencies in measurement protocols, and varying levels of narrative detail, underscore the need for dataset-specific evaluations. 

While the above mentioned challenges are key to translate the TFN modelling approach reliably across clinical domains, our work opens several promising avenues for future research. First, more data modalities e.g., images, omics data, wearable data, etc. can also be integrated, facilitating more informative representations. Second, personalised modelling could adapt the Nexus and prediction mechanisms to individual patients, thus facilitate precise, patient-specific predictions. Third, the learnt representations provide a foundation to develop digital twins or digital patient models that simulate disease trajectories and treatment responses. Fourth, mechanisms to update models with new patient data may facilitate longitudinal pattern learning along clinical workflows. Finally, supplemental causality-aware modelling~\cite{welch2024identifiability} could further improve interpretability and trustworthiness in decision-support. 

%\subsection{Conclusion}
In conclusion, TFN is a task-agnostic multi-modal model that contextualises and integrates irregular time series and unstructure clinical text from EHRs. We showed that the multi-modal representations learnt by TFN were clinically meaningful in post KTx care, capturing long-term temporal dynamics and producing attributions that aligned with clinical reasoning across three key outcomes: graft loss, graft rejection, and mortality. TFN consistently outperformed unimodal baselines and state-of-the-art clinical decision support model. Beyond post KTx care, the proposed modelling approach has the potential to be applicable in various clinical domains that are characterised by heterogeneous data sources, irregular longitudinal measurements, and rich narrative documentation.

\section{Methods}\label{sec4}

\subsection{Encoding structured data}
Here we detail method for encoding structured data~(time series and static patient data), the mechanism used to deal with irregular and sparse time series data, and handling of long-term temporal dependencies in patient trajectories.

\subsubsection{TM-LSTM for time series data}
%The TM-LSTM can model varying time intervals between clinical measurements. The TM-LSTM can selectively discount short-term information based on elapsed time while preserving long-term dependencies. LSTM cells treat all temporal sequences as if observations occur at regular intervals, which fails to properly account for the clinical reality where visits may be separated by days, weeks, or months. 
In \textit{TM-LSTM} module, we handled irregular sample intervals through a memory decomposition approach. First, we separated the previous cell memory~($C_{t-1}$) into short-term~($C_{t-1}^S$) and long-term~($C_{t-1}^T$) components:

\begin{equation}
C_{t-1}^S = \tanh(W_{decomp} \cdot C_{t-1} + b_{decomp})
\end{equation}
\begin{equation}
C_{t-1}^T = C_{t-1} - C_{t-1}^S
\end{equation}
The memory decomposition discounted only the short-term memory based on elapsed time, using a decay function:
\begin{equation}
C^*_{t-1} = C^T_{t-1} + C^S_{t-1} \odot e^{(-W_{decay} \cdot \Delta t + b_{decay})},
\end{equation}
where $\Delta t$ is the time elapsed since the previous observation. \\
The adjusted memory $C_{t-1}^*$ was put into the standard LSTM gates. The idea behind the memory decomposition is to appropriately model the diminishing relevance of past events based on the temporal distance by discounting short-term effects proportionally to time gaps. The learnt parameters $W_{decay}$ and $b_{decay}$ allows the model to adaptively determine feature-specific decay rates, capturing the varying clinical significance of different measurements over time~\cite{tlstm_orig}.

In \textit{TM-LSTM}, we also employed a feature-level masking mechanism to handle missing values in the time series data. For each time step, we created a binary mask $M \in {0,1}^{B \times T \times F}$ ,where $B$ is the batch size, $T$ is the sequence length, and $F$ is the number of features. The idea behind the mask is to indicate which feature values are observed~($M[b,t,f] = 1$) or missing~($M[b,t,f] = 0$) at a given time point. The masking strategy enables the model to distinguish between observed values and missing data points. During model training and inference, the state updates were selectively applied based on the observed features, while missing features were ignored. The selective updating mechanism preserves the original temporal structure of the data without requiring imputation, mitigating potential biases~\cite{survey3}.

We then implemented a multi-head self-attention mechanism over the \textit{TM-LSTM} outputs to capture the variable importance of different time points in the patient's history. Given the hidden states $H = {h_1, h_2, ..., h_T}$, we computed:
\begin{equation}
\text{Attention}~(Q, K, V) = \text{softmax}\left(\frac{QK^T}{\sqrt{d_k}}\right)V,
\end{equation}
where $Q = W_Q H$, $K = W_K H$, and $V = W_V H$ are linear projections of the hidden states. The attention was applied with a causal mask to ensure that predictions at time $t$ only depend on observations up to time $t$. Additionally, we applied layer normalisation and feed-forward networks with residual connections to prevent overfitting. Attention allows the model to focus selectively on time-steps that most are most influential for future clinical outcomes, rather than treating all historical observations equally. 
Furthermore, it improves the model's stability for long sequences~\cite{attain}.

\subsubsection{Static feature integration}
We incorporated static features using a dedicated static encoder. Numerical features were processed by a linear layer and categorical features were transformed using embedding layers. Embedding layers learnt small dense vector representations of nominal variables including blood groups or underlying diseases, which captured semantic relationships between categories that would be lost in simple one-hot encoding.

The resulting static representations were concatenated and integrated with time series data through a variable-wise gating mechanism that selectively updated the \textit{TM-LSTM's} hidden state at each time step. Variable-wise gating allowed the model to control how much static information influenced temporal processing, facilitating patient-specific interpretations of the same clinical time series. %For instance, the model could learn that identical vital sign patterns carry different clinical significance for elderly patients versus younger ones. 
A residual connection was used to increase training stability.

\subsection{Encoding unstructured data}
%For encoding clinical notes, we utilised med-gte-hybrid-de, a finetuned sentence transformer for German medical texts. 
%Our text encoder module, \textit{Med-GTE-hybrid-de} is based on GTE-large, a general pre-trained sentence transformer. 
We fine-tuned \textit{Med-GTE-hybrid-de} through a dual-approach fine-tuning strategy using 20000 randomly sampled German clinical sentences from our dataset. The model was trained through two distinct self-supervised stages: contrastive learning followed by denoising autoencoder training. Contrastive learning encourages the model to bring semantically similar sentences closer in the embedding space while pushing dissimilar ones apart, improving model's ability to distinguish clinically meaningful concepts. Denoising encourages the model to learn robust representations that capture the underlying meaning of clinical text despite noise, inconsistent terminology, or other variations typical in medical records. The combined approach enables the model to generate embeddings that effectively support various downstream clinical tasks, including mortality prediction and disease prognosis~\cite{medgte}.

In the contrastive learning, we processed each input sentence twice through the encoder with different dropout masks, creating two slightly different embeddings that were treated as positive pairs. Positive pairs were trained to have high similarity in the embedding space, while all other sentence embeddings in the same batch were treated as negative examples and pushed apart.

In denoising autoencoder component, we created corrupted sentences~(e.g., by randomly masking or introducing noise) and then trained the model to reconstruct original sentences from corrupted versions. 

\subsection{Integration of modalities via cross-attention}
We integrated time series and textual representations using a cross-attention mechanism. The cross-attention mechanism enables each clinical time point to retrieve the most relevant contextual information from the accompanying clinical notes. As illustrated in Fig.~\ref{overview}, queries $Q$ were computed from the \textit{TM-LSTM} hidden states 
$\mathbf{H}_{ts} = \{ \mathbf{h}^t_1, \ldots, \mathbf{h}^t_n \}$, 
while keys $K$ and values $V$ were obtained from linearly projected clinical text embeddings 
$\mathbf{H}_{\text{text}} = \{ \mathbf{e}^t_1, \ldots, \mathbf{e}^t_m \}$. 
Formally, \(Q = \mathbf{H}_{ts} W_Q,\) \(K = \mathbf{H}_{\text{text}} W_K,\) \(V = \mathbf{H}_{\text{text}} W_V\), where $W_Q$, $W_K$, and $W_V$ are learnable projection matrices.

We computed the interactions between modalities using the scaled dot-product formulation:
\begin{equation}
\mathrm{Attention}~(Q,K,V)
= \mathrm{Softmax}\!\left( \frac{QK^\top}{\sqrt{d_k}} + M \right)V,
\end{equation}
where $d_k$ is the key dimensionality and $M$ is a causal mask that restricts each time point $t_i$ to attend only to notes documented at or before $t_i$, thereby preventing information leakage from future clinical narratives, ensuring temporal consistency across modalities.

The attention output $X$ was subsequently passed through a feed-forward integration layer: Z = $\Sigma(X)$, where $\Sigma(\cdot)$ denotes a learnable transformation that fused the attended textual information with the temporal embeddings. The resulting representation was a multi-modal shared embedding space, termed as the Nexus.

\subsection{Model training}
In our training approach, we leveraged the temporal nature of patient data to create a self-supervised learning scheme. For each patient, we generated multiple training samples by incrementally extending the historical window of clinical measurements and predicting future values.
Given a patient's time series data $X = \{x_1, x_2, ..., x_t\}$ where $x_t$ represents a data point at time $t$, we created training samples by using increasing history lengths to predict future data points. For each time point $t$, we used the sequence $\{x_1, x_2, ..., x_t\}$ as input to predict $H$ future steps where $H$ is called the prediction horizon. 
The self-supervised approach created multiple samples per patient while maintaining the sequential nature of the data, enabling the model to learn temporal patterns at various stages of patient history. 

\subsubsection{Loss function}
The TFN model was trained using a composite loss function that combined three components:

\begin{equation}
\mathcal{L}_{total} = \mathcal{L}_{recon} + \lambda_{decorr} \cdot \mathcal{L}_{decorr} + \lambda_{disent} \cdot \mathcal{L}_{disent},
\end{equation}

where $\lambda_{decorr}$ and $\lambda_{disent}$ are hyperparameters controlling the contribution of each loss term.

\textbf{Reconstruction Loss:} This term quantified the model's ability to predict future time steps accurately. The loss function:

\begin{equation}
\mathcal{L}_{recon} = \frac{1}{H} \sum\limits_{h=1}^{H} (x_{t+h} - \hat{x}_{t+h})^2,
\end{equation}

where $\hat{x}_{t+h}$ is the model's prediction for time step $t+h$.

\textbf{Decorrelation loss:} This regularisation term encouraged minimal correlation between latent dimensions by specifically targetting off-diagonal elements of the covariance matrix, encouraging independence between different latent dimensions and reducing redundancy in the learnt representations. The loss function:

\begin{equation}
\mathcal{L}_{decorr} = \sum_{i \neq j} (Cov(Z)_{i,j})^2,
\end{equation}

where $Z \in \mathbb{R}^{N \times K}$ is the mean-centered matrix of latent representations for a batch of $N$ samples with $K$ latent dimensions, and $Cov(Z)$ is the covariance matrix.

\textbf{Disentanglement loss:} This term encouraged separation of features in the latent space through feature decoders with sparse soft masks. The loss function:

\begin{equation}
\mathcal{L}_{disent} = \frac{1}{D} \sum_{d=1}^{D}({{f_d} - \hat{f}_d}(Z))^2 + \alpha \|{M_d}\|_1,
\end{equation}

where $\hat{f}_d(Z)$ is the prediction of feature $d$ from the latent representation using a sparse mask $M_d$, $f_d$ is the ground truth feature, $\|\cdot\|_1$ is the L1 norm promoting sparsity, and $\alpha$ is a hyperparameter controlling mask sparsity.

\subsubsection{Implementation details}
The TFN model was trained using five-fold cross-validation, with 80\% of patients allocated to training and 20\% to evaluation in each fold. We reported mean performance metrics across folds along with standard deviations. We created training samples with a minimum of 10 data points and used a prediction horizon of 10. See Supplemenatry Appendix B for hyperparameter choices.  

\subsection{Evaluation}
Here, we detail evaluation techniques utilised for predictive performance analysis, assessment of learnt latent structure, as well as calibration and explaniability of our TFN model.

\subsubsection{Downstream predictive performance}
%We evaluated our model on three critical clinical prediction tasks: \textbf{graft loss prediction}, \textbf{graft rejection prediction}, and \textbf{mortality prediction}.
For each of the three clinical prediction tasks~(graft loss, graft rejection and mortality), we predicted outcomes within multiple time windows. At each time point $t$, the model processed the patient's history $\{x_1, x_2, ..., x_t\}$ and generated the patient representation. Each representation was put into a small binary classifier. The classifier was trained using oversampling to deal with the imbalanced datasets. 

We assessed prediction performance using the ROC curve, plotting sensitivity against specificity across classification thresholds. The AUC quantifies overall discriminative performance independent of threshold selection. Additionally, we reported sensitivity and specificity at a 0.5 threshold. In practice, the threshold can be adjusted based on clinical priorities to favour either metric~\cite{sens}. Sensitivity measures the correct identification of patients experiencing adverse outcomes~(true positive rate), while specificity measures the accurate identification of event-free patients~(true negative rate). In transplantation medicine, high sensitivity ensures identification of patients requiring intensified monitoring, whereas high specificity prevents unnecessary interventions in low-risk populations~\cite{ckd_pred}.

Evaluation was conducted on a per-patient basis, with metrics calculated by aggregating predictions across all available time points for each patient. 

\subsubsection{Assessment of structure of the Nexus}
%We evaluated the effect of our decorrelation and disentanglement penalty on the Nexus of TFN model.  
We applied the DCI metrics~\cite{dci} to assess the structure of the Nexus quantitatively. The DCI metrics directly evaluate the structure of the representation space, with higher scores indicating better disentanglement. 

The components of DCI framework include:\\
\textbf{Disentanglement}: The disentanglement metric quantifies how uniquely each latent dimension corresponds to a specific clinical factor. It is computed by training factor-predictor models and examining the extent to which each latent dimension contributes to each prediction. High scores indicate that individual dimensions predominantly influence a single clinical factor, reflecting a structured representation.\\
\textbf{Completeness}: The completeness metric measures whether each clinical feature is encoded in a single dimension rather than spread across many. It is computed by analysing importance weights when predicting input features from latent variables. High values indicate that each feature's information is concentrated in a few dimensions.\\
\textbf{Informativeness}: The informativeness metric measures how accurately original features can be reconstructed from the representation. It is calculated as the average prediction performance when predicting input features from latent variables.

\subsubsection{Model reliability}
%While the discriminative ability of our model is crucial, the reliability of its probabilities is equally important in clinical settings. In clinical practice, there is a meaningful difference between a model that predicts an outcome with 51\% confidence versus 99\% confidence. However, confidence calibration is not directly captured by discrimination metrics. Especially neural networks tend to be poorly calibrated, often producing overconfident probability estimates that do not match empirical outcomes~\cite{calibration}. 
We evaluated our model's calibration using calibration curves, which visualised the alignment between predicted probabilities and observed frequencies of events. We then applied Platt scaling to improve model calibration. Platt scaling is a post-processing technique that uses logistic regression on the model's outputs to transform raw predictions into well-calibrated probabilities, without changing the model's discriminative performance~\cite{platt}. We utilised Brier score~\cite{redelmeier1991assessing} to quantify the accuracy of probabilistic predictions by measuring the mean squared difference between predicted probabilities and actual binary outcomes. A low Brier score indicates good calibration, meaning predicted probabilities closely reflect observed probabilities. 

\subsubsection{Explainability study}
We leveraged SHAP values to investigate the importance of the features driving our TFN model's predictions~(graft loss, rejection, and mortality). SHAP allows us to identify the features on which the model places the greatest emphasis, providing transparency into the prediction process. For clinicians, transparency enhances model trustworthiness by making the reasoning behind predictions interpretable~\cite{reliable}.

To evaluate whether the model's feature importance aligned with clinical expertise, we asked a nephrologist with over 30 years of experience in transplantation medicine to rank features. We compared the expert ratings with the TFN feature importance rankings using Spearman's rank correlation coefficient. The quantitative comparison allowed us to assess the degree of alignment between algorithmic feature importance and expert clinical judgment. Additionally, we conducted a qualitative analysis of the highest-ranked features to determine whether the TFN model emphasised clinically meaningful indicators in decision-making process.

\section*{Declarations}

\subsection*{Ethics approval and consent to participate}
This work is part of the smartNTx+ study, which was approved by the ethics committee of University Hospital Erlangen~(23-300-B). 
The NephroCAGE data collection was approved by the ethics committee of Charité University Medical Center Berlin~(EA4/104/21). All methods were performed in accordance with the relevant guidelines and regulations. Available data were pseudonymised, retrospective data obtained from patients, who gave their informed consent for the retrospective scientific analysis of their data prior to their transplantation. All data was handled in accordance with the corresponding data protection regulations, i.e., the European General Data Protection Regulation~(GDPR). 

\subsection*{Availability of data and materials}
\subsubsection*{Data availability statement} 
The NephroCAGE dataset is not publicly available. However, detailed information on the data dictionary can be made available from the authors on reasonable request.
 
\subsubsection*{Code availability}
To facilitate reproducibility and benefit other studies the source code has been shared on: \url{https://github.com/intelligentembeddedsystemslab/TFN_Temporal-Fusion-Nexus}.

\subsection*{Competing interests}
The authors declare no conflict of interest.

\subsection*{Funding}
This work recieved funding by the German G-BA, in Project SmartNTx, grant number 01NVF21116. NephroCAGE was funded by German Ministry for Economic Affairs and Climate~(01MJ21002)

\subsection*{Authors' contributions}
A.K. conceptualised the work and prepared the data. S.R. performed model training. S.R. and A.K. designed and performed model evaluations, ablations and other analyses. A.K. drafted the initial manuscript. M.C. and O.A. provided research guidance, critically analysed results, and contributed substantially to the manuscript editing. A.K. supervised the project and administered the execution. M.N. contributed to feature selection for modelling. M.-P.S., M.N., A.R., F.H., and B.O. are authors of NephroCAGE and were involved in data collection and dataset creation. M.-P.S., R.R., L.P, K.B., and M.S. contributed to the internal manuscript review. 

\begin{appendices}

\end{appendices}

\bibliographystyle{plain}
\bibliography{bibliography}

\end{document}